\documentclass[preprint,12pt,authoryear]{elsarticle}

\usepackage[utf8]{inputenc}

\usepackage{amssymb}
\usepackage{amsmath}

\usepackage{algorithm}
\usepackage{algpseudocode}
\usepackage{setspace}
\usepackage{graphicx}
\usepackage{subcaption}

\journal{Neural Networks}

\renewcommand{\cite}{\citep}

\begin{document}

\begin{frontmatter}



\title{MetalGAN: Multi-Domain Label-Less\\
Image Synthesis Using cGANs and Meta-Learning}


\author{Tomaso Fontanini}
\ead{tomaso.fontanini@studenti.unipr.it}
\author{Eleonora Iotti}
\author{Luca Donati}
\author{Andrea Prati}
\address{IMP Lab, Department of Engineering and Architecture, University of Parma, Parco Area delle Scienze, 181/A, 43124 Parma}

\begin{abstract}
Image synthesis is currently one of the most addressed image processing topic in computer vision and deep learning fields of study. Researchers have tackled this problem focusing their efforts on its several challenging problems, e.g. image quality and size, domain and pose changing, architecture of the networks, and so on.
Above all, producing images belonging to different domains by using a single architecture is a very relevant goal for image generation.
In fact, a single multi-domain network would allow greater flexibility and robustness in the image synthesis task than other approaches.
This paper proposes a novel architecture and a training algorithm, which are able to produce multi-domain outputs using a single network. A small portion of a dataset is intentionally used, and there are no hard-coded labels (or classes).
This is achieved by combining a conditional Generative Adversarial Network (cGAN) for image generation and a Meta-Learning algorithm for domain switch, and we called our approach MetalGAN.
The approach has proved to be appropriate for solving the multi-domain problem and it is validated on facial attribute transfer, using CelebA dataset.
\end{abstract}




\end{frontmatter}


\section{Introduction}\label{sec:intro}

Image generation or synthesis consists in the act of producing a novel image---representing a subject of interest or whatever else---from an input that could be a random noise matrix, another (real) image, or a combination of these two possibilities, eventually put beside a label or a condition that somehow controls the output.
The required output should belong to a specific domain, or it should have been obtained following a precise style. Conversely, in some cases the image domain or style could be not decided \emph{a-priori}, and the developed system should perform multi-domain image generation. 

Since the recent advances of deep learning training techniques and architectures on image generation, the image synthesis task has become more and more accessible and understood.
Examples of such techniques are the use of \emph{Generative Adversarial Networks} (\emph{GANs}) (in particular \emph{Deep Convolutional} GANs, called DCGANs) and \emph{conditional GANs} (\emph{cGANs}), which should have different architectures, such as \emph{U-Nets} and \emph{StyleGAN}, and different training methods, like \emph{pix2pix}, \emph{cycleGAN}, and so on. A high number of specific approaches were developed to face the aforementioned variety of image generation tasks, leading to a vast literature for each specific sub-problem. A brief outline of the major methodologies are reported in the next section.

This paper focuses on the specific problem of image-to-image translation.
Image-to-image translation is the act of transforming an arbitrary image in another, more useful, representation of the same data. Image colorization, semantic segmentation, style transfer are examples of image-to-image translations.
In particular, this work approaches the task of transforming the input image over a range of so-called \textit{domains}, i.e. recognizable sets of similar images which share common characteristics. 
One of the scope of this work is to develop a single architecture that is able to handle many different domains, i.e. a multi-domain image-to-image generator system. More specifically, in a system like that, the architecture is required to learn multiple mapping functions, that is one for each domain. In our case, the multiple domains are represented by different facial attributes like ``blond hair'' or``pale skin''  and the mapping functions have the objective to apply these attributes on any face passing through the architecture.
Moreover, the idea is to make the model capable to learn new, unseen, domains by using few images for each new domain, in order to gain a great flexibility and generalization capability of the proposed architecture and to tackle also those applications or domains where there is scarcity of available data.
Hence, the topic covered in this work is \emph{multi-domain image-to-image translation}, deeply entranced with the concept of domain adaptation.

One interesting take on this topic is that many of these image-to-image transformations are linked by a common way of working. For example, changing hair color, e.g. switching the domain to the new one of ``people with blond hair'', needs to correctly segment hair in the same way of changing the domain in ``people with black hair''. Similarly, changing a face into its older version and add glasses to a face both need to correctly locate the subject's eyes. 
Yet, for long time, a single neural network for each of these tasks had to be created, even if the tasks were quite similar. A solution to this kind of problem has been proposed with StarGAN~\cite{choi2018stargan}, which had the intuition of bringing together multiple image-to-image transformations in the same network architecture.

Another observation is that most of the existing approaches to image-to-image translations perform a full training with input-output examples of images, to achieve high quality results. An evident drawback of this approach is that they need very large datasets to be trained.
Dataset could be labeled or unlabeled, but usually the domain switch is controlled by a conditioning label, which indicates the target domain to transform the image to. Hence, image-to-image translation often requires a lot of labeled images, where the label denotes the domain(s). It is worth noting that an image could have more than one label: this is the reason for not addressing labels as classes.

Regarding the adaptation of the model to new domains with few images, a similar issue is the few-shot learning problem.
Few-shots problems are often addressed by using \emph{meta-learning} techniques, thanks to their ability to switch among a distribution of tasks during training. Training in a meta-learning settings means creating a learning system that includes another learning sub-system: the sub-system trains a model (such as a neural network) on a single task sampled from the distribution, and the meta-learning system trains the sub-system, thus adapting the model to all tasks.
Meta-learning methods have proved to be successful in classification and regression scenarios, but there are still few papers~\cite{liu2019few, zhang2018metagan}
in the field of image generation.

Linked to domain adaptation, another known problem of traditional training settings is that once a new set of tasks emerges, e.g. a new domain is added to the target (or desired) outputs, a full retraining of the whole system is needed. This happens even if the new task is similar to tasks that the network has already learned.
The full re-training includes incorporating the new domain in the input examples and also it often needs architecture changes.

The main proposal of this article, and its principal contributions are:
\begin{itemize}
	\item a system that consists in a single cGAN (i.e., two networks, a generator and a discriminator) performing image-to-image translation, trained on multiple domains;
	\item both networks do not contain any reference to the label or domain of the input or output (\textit{label-less}) therefore allowing a much more flexible architecture;
	\item the system is able to switch task with just few examples of a new, unseen domain, by means of a meta-learning training algorithm. This was impossible in previous architectures representing a great limitation;
	\item the system uses knowledge accumulated at training time with well-known, largely represented classes, to easily learn new, unknown tasks in few iterations.
\end{itemize}

Taking into account all these contributions and the main proposed idea to fuse together meta-learning and GAN, we named our approach \textit{MetalGAN}.

The paper is organized as follows. Section \ref{sec:related} presents an extensive evaluation of the state-of-art. Section \ref{sec:metalgan} introduces a complete overview of the system: main idea and notations, architecture of the network and algorithm. Section \ref{sec:experiments} describes the experimental results. Finally, Section \ref{sec:conclusions} presents our conclusions.

\section{Related Work}
\label{sec:related}

\paragraph{\bf Image-to-image translation}
The main topic of this paper, i.e. image-to-image translation, has become a hot topic in machine learning researcher community after the introduction of encoder-decoder networks like U-Nets \cite{ronneberger2015u}, \emph{Fully Convolutional Neural networks} (\emph{FCN}) \cite{long2015fully} and conditional GANs \cite{mirza2014conditional}.
The GAN approach to image synthesis has proven an unprecedented quality of output results, reaching photorealism in many domains, such as face synthesis. 
While traditional GANs~\cite{goodfellow2014generative} generate images from noise, conditional GANs (cGANs)~\cite{mirza2014conditional} in their many variations are able to generate images from labels or other input images, or both. To this extent, cGANs are often used to perform lots of different image-to-image translation tasks like producing sketch colorization and texture generation~\cite{sangkloy2017scribbler,xian2018texturegan}, super-resolution of images~\cite{ledig2017photo} or to generate a photo-realistic image from a semantic label map~\cite{wang2018high, park2019semantic}.
cGANs can be trained in both a paired \cite{isola2017image, zhu2017toward} or unpaired way \cite{zhu2017unpaired, almahairi2018augmented, kim2017learning}. 

In our approach, we use cGANs without a paired dataset with only input image but label-less, in order to maintain a great generalization capability of the generator network. Moreover, we introduced skip connections in the generator network, as in U-Nets.

\paragraph{\bf Multi-domain image-to-image translation} A common trait of most of the image-to-image methods is that they are only able to produce outputs belonging to a single domain or class. 
Regarding multi-domain facial attributes transfer, our main work of reference is StarGAN~\cite{choi2018stargan}, though there exist other relevant works like \cite{he2019attgan,xiao2017dna, kim2017unsupervised}. 
StarGAN proposes an unified method for multi-domain image-to-image translation. 
It achieves great results in image synthesis taking strength from the multiple domain adaptations and it learns multiple domains at the same time using only one underlying representation. The main differences between StarGAN and the proposed method are: in our approach networks do not use labels information (while StarGAN do); our training method relies on a small number of images per-iteration; and also a few-shot-like approach is employed when dealing with new domains during inference.

\paragraph{\bf Few-shots learning}
Few-shot problems are usually tackled with meta-learning techniques, since recent results show great performance of meta-learners on typical few-shot datasets and learning settings.
There are many types of meta-learners. Some learn how to parameterize the optimizer of the network \cite{hochreiter2001learning,ravi2016optimization}, while others use a network as optimizer  \cite{li2017learning,andrychowicz2016learning, wichrowska2017learned}. Furthermore, using a recurrent neural network trained on the episodes of a set of task is one of the most general approach~\cite{santoro2016meta, mishra2017simple, duan2016rl, wang1611learning}.
For our work, the most relevant meta-learners are the ones based on hyper-parameterized gradient descent such as Reptile~\cite{nichol2018first} and MAML~\cite{finn2017model}. In fact, we use the Reptile algorithm applied to a generation problem, where Reptile tasks are identified with our domains. Reptile was already used in combination with GANs in \cite{clouatre2019figr} in order to generate very simple black and white images (such as MNIST digits) or in \cite{zhang2018metagan} that introduced an adversarial discriminator, conditioned on tasks.

Regarding few-shots image-to-image translation, a new method was recently introduced in \cite{liu2019few}, coupling an adversarial training scheme with a novel network design. Unlike our method, it does not use meta-learning and does not act as a proper domain transfer algorithm, but rather as a style transfer one: for example, in the case of face image translation task, the translation output maintains the pose of the input content image, but the appearance is similar to the the faces of the target person.

\section{Overview of the System}
\label{sec:metalgan}

\subsection{Idea and Notations}
\label{sec:idea}

As briefly outlined in the introduction, there are some key points from which our work originates, namely, the need of a \emph{few-shots} setting, the use of a \emph{single} GAN architecture, the \emph{absence of labels}, and the \emph{multi-domain} adaptation.
All these key points require a proper definition.

Starting from the most potentially ambiguous definition, we call a ``domain'' a set of images which share a well-defined common characteristic, clearly recognizable by using a single label or keyword: for example, ``black-hair'' in a dataset of faces denotes the domain of people with a black hair color. Given the example above, it is also clear that the type of dataset is also important: if the dataset contains both dogs and cats images, ``black-hair'' should have another meaning; if it contains only landscapes, ``black-hair'' should have no meaning at all. 
Moreover, domains are not mutually exclusive, rather they could intersect each other.

Closely related to the concept of domain, there is the concept of "label". Usually, when approaching multi-domains problems, labels are employed to identify which domains a certain image belongs to. This helps the networks in detecting a target domain and thus generating images belonging to such a domain. In our case, \textit{label-less} means that the domain of the input and the target images have to be inferred from other information.

In a classic few-shots classification setting, from which we borrow the notations, there are $n$ classes $\{c_1, c_2, \dots, c_n\}$ and a certain number $k$ of input examples per-class, e.g. $\{x_1, x_2, \dots, x_{k_i}\}$ are the input of the $i$-th class $c_i$.
During training only $N$ classes per iteration are used over the total number of classes $n$, and for each of these $N$ classes only $K$ input examples over the total number of examples of a class are used, where $K << k$. Then, the trained $N$-classifier has to classify a new example of a random class $\tilde{c}$.
In our case, domains are treated as classes, and the generator-discriminator (from now on, called $G$ and $D$) networks are trained on a single domain per meta-iteration ($N = 1$), in order to make $G$ and $D$ able to learn the domain they are working on, without labels.
The number $K$ of examples per domain varies according to the type of experiment performed (see Section~\ref{sec:experiments}), but we choose to perform an almost full training and a few-shot inference to allow $G$ to learn adequately the reconstruction of images, and then to switch domain.
The architecture of our multi-domain GAN is detailed in Section~\ref{sec:architecture}.

Finally, in the meta-learning nomenclature, we defined a \emph{task} as a group of $K$ images that belong to the same domain, used for the inner-iteration of the algorithm, explained in detail in Section~\ref{sec:algorithm}.

Our approach uses a single GAN on different tasks.
This forces the underlying weights structure of both $G$ and $D$ networks to learn a general yet effective representation for describing all tasks.
$G$ and $D$ networks are conditioned with the use of a meta-learning algorithm, on each task/domain.
Other approaches, like StarGAN, instead, needs target labels that condition the output for both $G$ and $D$ networks.
In detail, the conditioning is implicitly provided by the task selection performed during meta-learning. 
For each meta-iteration, a single task is selected, and the network is trained on that single task for a number of internal iterations.
In the next meta-iteration the training is performed on another, different but related task. 
With this training algorithm the network learns, meta-iteration after meta-iteration, a representation that is good (but not optimal) in performing all tasks, and just needs a little final push (few epochs of training) to be moved in the direction of the target task.


\subsection{Architecture of the Network}
\label{sec:architecture}

One of the strengths of our proposal is that it completely removes the need of providing specific labels for the data, because the network does not use one-hot labels or similar. If data are already labeled, labels are only useful in the preprocessing phase for dividing into domains the dataset, since the main algorithm works task-by-task. It is worth noting that such domains could overlaps.
On the other hand, unlabeled data has to be clustered into domains, a passage that can be completely automated (in contrast with manual labeling), but using a clustering method, domains does not overlap. A clusterization followed by a meta-learning approach is shown in a preliminary work on colorization~\cite{fontanini2019metalgan}.

\begin{figure}[]
	\centering
	\includegraphics[width=\textwidth]{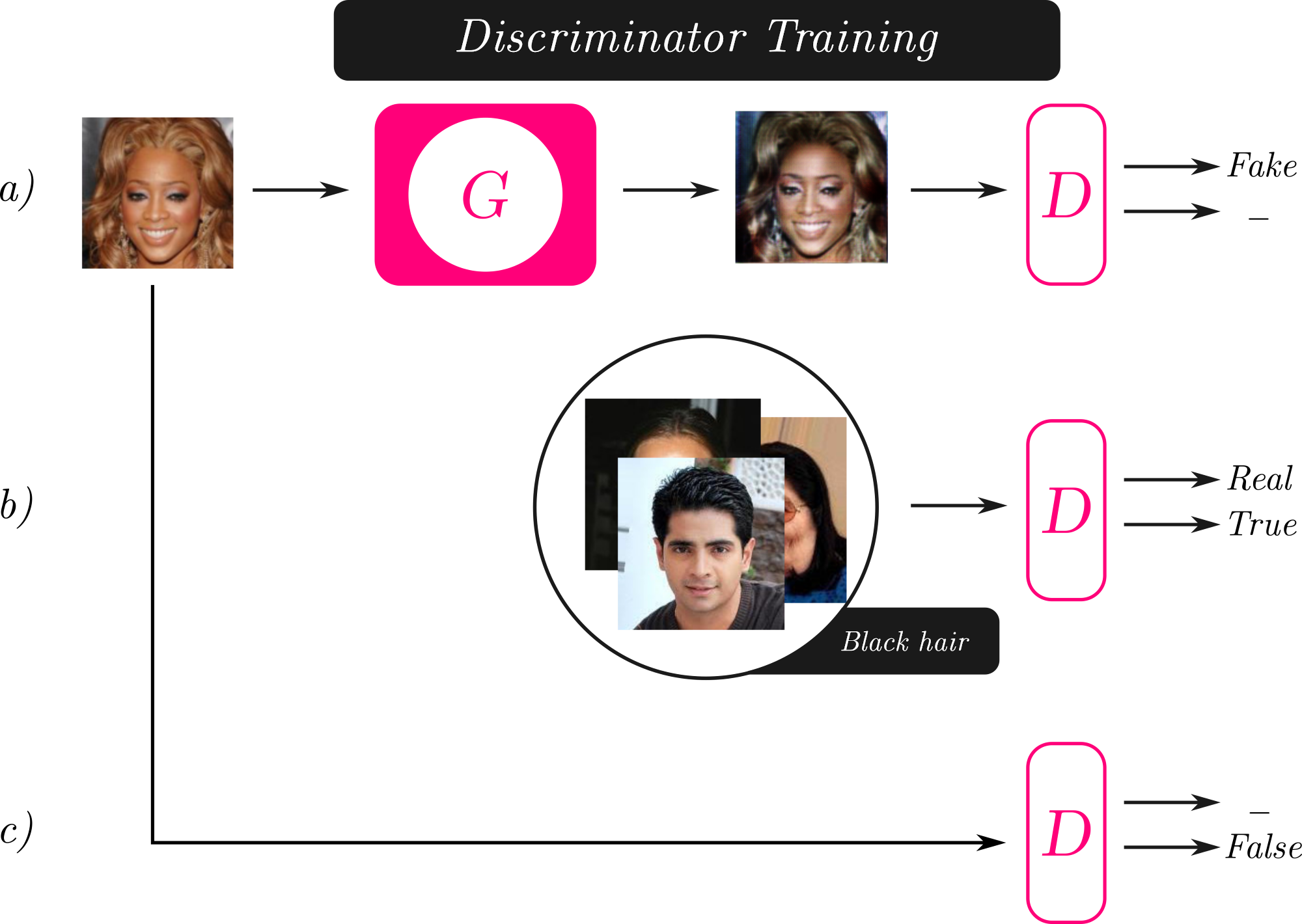}\\
	\vspace{1cm}
	\includegraphics[height=0.4\textwidth]{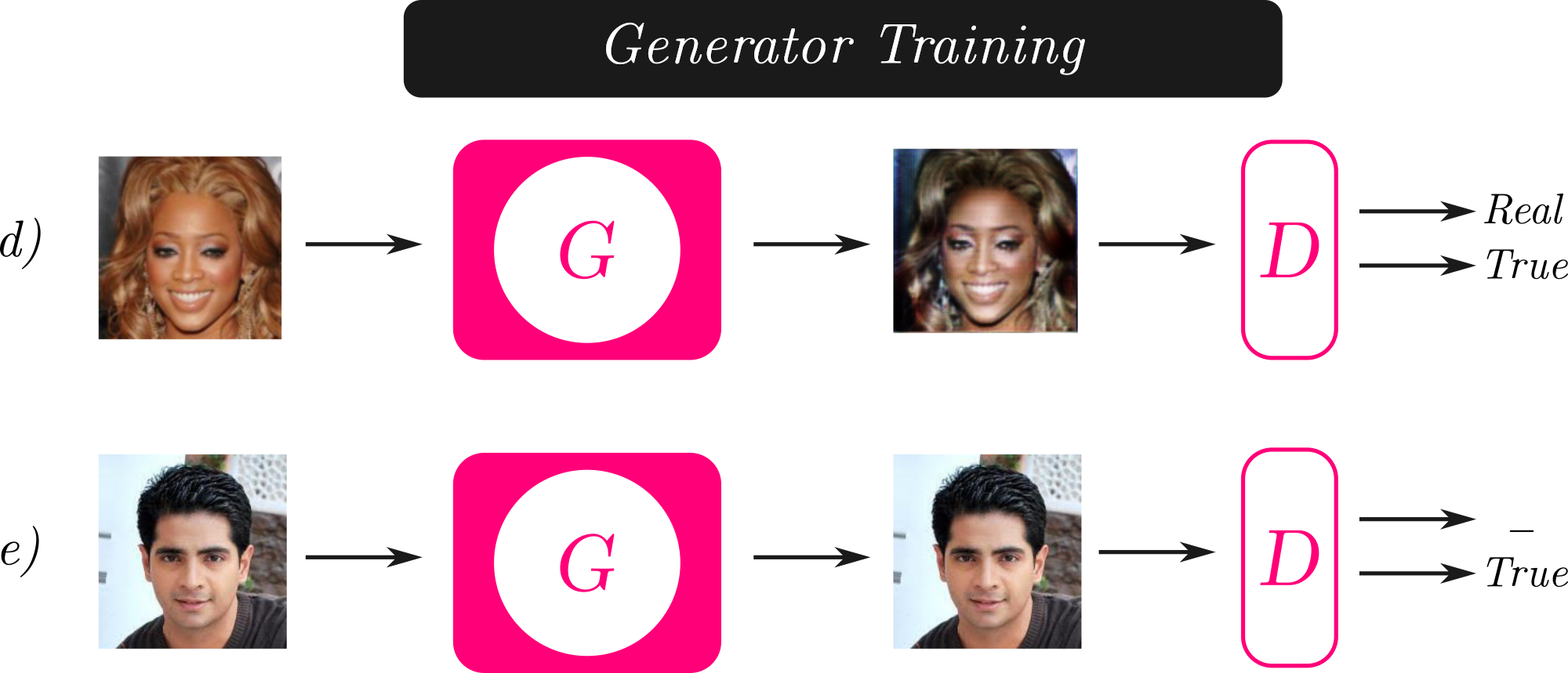}
	\caption{Complete network architecture. First, the discriminator is trained to distinguish between fake images (a) and real ones (b) and between images belonging to the current domain (b) and images that do not belong to it (c). Then, the generator is trained to fool the discriminator by labeling its outputs as real and as belonging to the current domain (d). Finally, the reconstruction step is executed and its results are labeled as part of the current domain (e).  }
	\label{fig:gd}
\end{figure}

Our system is composed by a single cGAN. In particular, since the objective is to generate the face of a person with only a bunch of new attributes, without changing the peculiar traits of the person itself and without using labels, we conditioned the cGAN with the input face, in order to maintain the identity of the person, and, at the same time, changing the target attributes. 

The generator network $G$ is the same as the StarGAN one with the addition of skip-connections (inspired by the classic U-Net), but input labels are removed.
The introduction of skip-connection in the generator architecture aims at enhancing the quality of the reconstruction; in other words, they are useful for keeping contents of the input image unchanged in the output image (for example, the face of a person remains the same despite the changing of hair color).

On the other side, the $D$ structure is the PatchGAN from pix2pix~\cite{isola2017image}. Since during each task the network tunes itself on a single domain, there is no need of a domain classification output for our discriminator. Instead, we choose to classify images both as real or fake and as belonging or not to the current domain. By doing so our discriminator has two outputs: $D_{adv}(x)$ and $D_{dom}(x)$, one for each probability distribution.

Finally, we define a set of losses in order to train our architecture.

\paragraph{\bf Adversarial Loss}
For the discriminator, we use an adversarial loss to distinguish between real and generated images:

\begin{equation}\label{eqn:adv_loss}
\mathcal{L}_{\mathrm{adv}}(D, G) = \mathbb{E}_{y \sim p_{\tau}} \left[ \log{D_{adv}(y)} \right] + \mathbb{E}_{x \sim p_{\mathrm{data}}} \left[ 1 - \log{D_{adv}(G(x))} \right],
\end{equation} 
where $y$ is sampled over a distribution of current task images $p_{\tau}$ (real samples), and $x$ over the distribution of the whole dataset $p_{\mathrm{data}}$ ($G(x)$ are the generated samples).

In particular, during each task, $D$ tries to classify if an image (or a batch of images) $y$ belongs or not to the current domain distribution $\tau$ (all images in the batch must belong to the domain). 
For example, if the current task is to produce people with blond hair, the discriminator has to determine if an image contains a person with blond hair or not, while in classic adversarial settings it should simply decide if the image contains a face or not.
The adversarial loss formula~\eqref{eqn:adv_loss} does not reflect explicitly this aspect, since the only difference is the nature of the input $y$: in our work, $y$ is not concatenated to any label.

\paragraph{\bf Domain Loss}
After we select a new task, during the training of the discriminator, we want images sampled from the current task to be classified as such and, on the other side, images sampled from the whole dataset to be classified as not belonging to the current task.

\begin{equation}\label{eqn:domain_loss_i}
\mathcal{L}'_{\mathrm{dom}}(D) = 
2 \cdot \mathbb{E}_{y \sim p_{\tau}} \left[ \log{D_{dom}(y)} \right] + \mathbb{E}_{x \sim p_{\mathrm{data}}} \left[ 1 - \log{D_{dom}(x)} \right],
\end{equation} 
where the multiplicative factor before $\mathbb{E}_{y \sim p_{\tau}} \left[ \log{D_{dom}(y)} \right]$ is motivated by the fact that we need to take into account that an image $x$ may also belong to the domain identified by the current task, since it is drawn from the whole data distribution. For this reason, the first part of the equation strongly reinforces the classification of examples of the target domain, while the second part weakly penalizes every domain (that is, also the target one).

Instead, during the generator training, the goal is that all the generated images, even the ones obtained from the reconstruction of the input, would be classified as belonging to the current task.

\begin{equation}\label{eqn:domain_loss_ii}
\mathcal{L}''_{\mathrm{dom}}(D, G) = 
\mathbb{E}_{x \sim p_{\mathrm{data}}} \left[ \log{D_{dom}(G(x))} \right] + \mathbb{E}_{y \sim p_{\tau}} \left[ \log{D_{dom}(G(y))} \right]
\end{equation} 

Adversarial and domain losses are visually described in Figure~\ref{fig:gd}.

\paragraph{\bf Reconstruction Loss}
This loss is crucial to  guarantee that the generator maintains the content information of the source image.
$G$ has to be already tuned on the current domain/task in the meta-learning training.
Since we completely removed the labels from our architecture and we tune the network on a new domain for each iteration, we cannot use a cycle consistency loss like in StarGAN, because $G$ is able to produce images of only one target domain each task.
The solution we choose to adopt is to apply the reconstruction loss on the images $y$ belonging to the target domain.
The reason is that if an image already belongs to the target domain, it should be left unchanged by $G$.
The equation for the reconstruction loss is as follows:

\begin{equation}
\mathcal{L}_{\mathrm{rec}}(G) = \mathbb{E}_{y \sim p_{\tau}}[|| G(y) - y ||_1],
\end{equation}
where $||\cdot||_1$ denotes the $L_1$ norm in the space of target images.

\paragraph{\bf Feature Matching Loss}
In order to regularize the training, we also include a feature matching loss following the work of \citep{wang2018high} and \citep{liu2019few}. Feature Loss stabilizes the training since it is required to the Generator to produce natural statistic at multiple scale. We extract features from the discriminator layers located before the prediction layer. This feature extractor is called $D_{\mathrm{feat}}$. The definition of the feature matching loss is as follows:

\begin{equation}
\mathcal{L}_{\mathrm{feat}}(D_{\mathrm{feat}}, G) = \mathbb{E}_{x,y \sim p_{\mathrm{data}}, p_{\tau}}[||D_{\mathrm{feat}}(G(x)) - D_{\mathrm{feat}}(y)||_1].
\end{equation}

\paragraph{\bf Full Objective} Finally, our full objective becomes:

\begin{equation}\label{eqn:full_obj_D}
\mathcal{L}_{D} = \mathcal{L}_{\mathrm{adv}} + \mathcal{L}'_{\mathrm{dom}},
\end{equation}

\begin{equation}\label{eqn:full_obj_G}
\mathcal{L}_{G} = w_{\mathrm{adv}} \mathcal{L}_{\mathrm{adv}} +
w_{\mathrm{dom}}\mathcal{L}''_{\mathrm{dom}} + 
w_{\mathrm{rec}}\mathcal{L}_{\mathrm{rec}} + 
w_{\mathrm{feat}}\mathcal{L}_{\mathrm{feat}},
\end{equation}
for the discriminator and generator, respectively. Furthermore, $w_{\mathrm{adv}}$, $w_{\mathrm{dom}}$, $w_{\mathrm{rec}}$, $w_{\mathrm{feat}}$ are the weights assigned to the loss functions. The discriminator loss functions do not have weights assigned since adversarial and domain losses should contribute equally to discriminator training to obtain balanced results. Weights choices are more properly discussed in Section~\ref{sec:experiments}.


\subsection{Algorithm}
\label{sec:algorithm}

Our approach relies on a meta-learning algorithm based on Reptile \cite{nichol2018first} and adapted to the image generation problem.

\begin{figure}[H]
	\vspace{0.5em}
	\centering
	\includegraphics[width=0.5\textwidth]{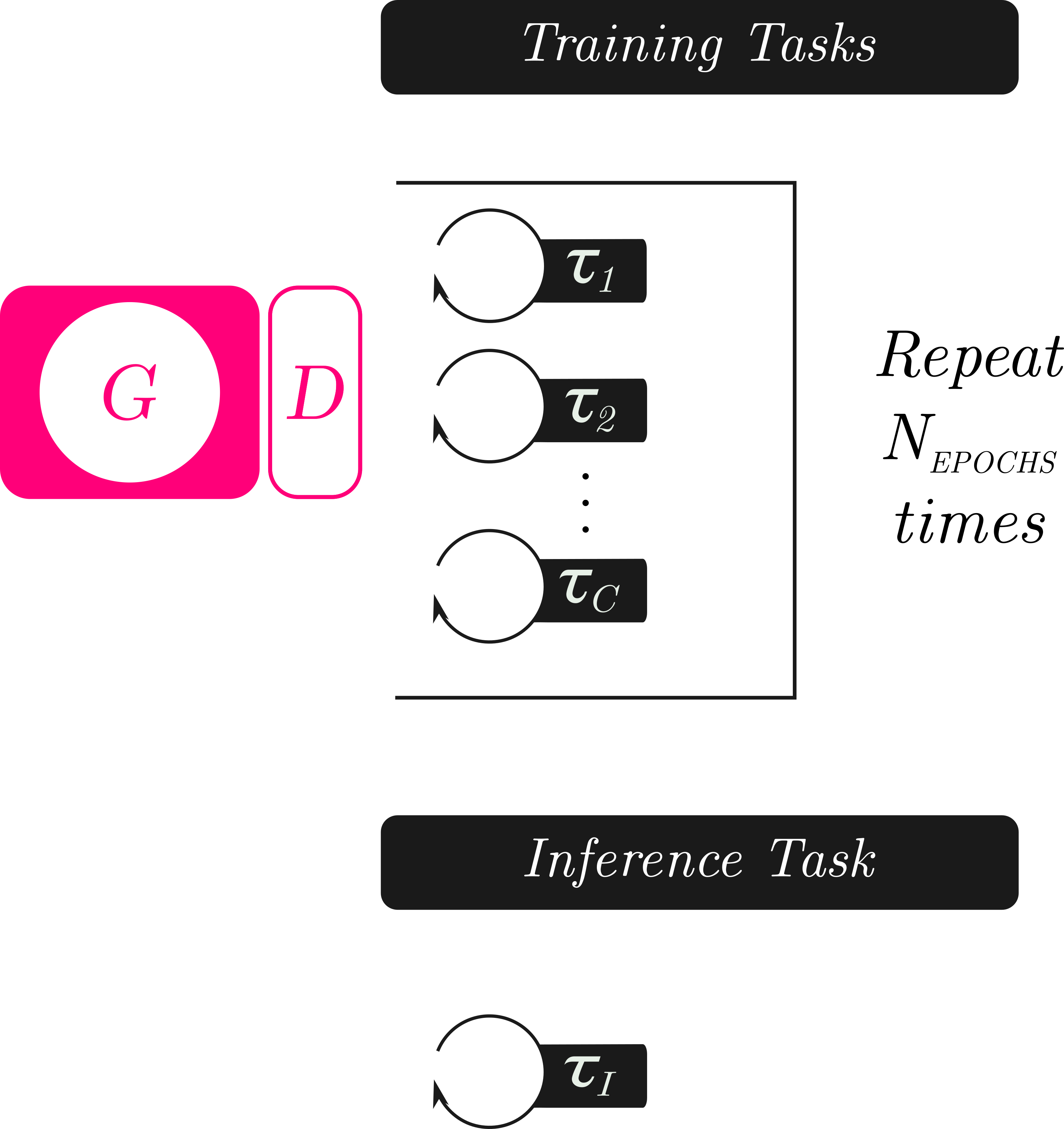}
	\caption{A full overview of the system: during the training phase, the network is trained for $N$ epochs on a set of tasks, and then, during the inference phase, a new unknown task, i.e. not present in the training phase, is selected and added to the network.}
	\label{fig:meta_train}
\end{figure}

\begin{figure}[H]
	\centering
	\includegraphics[width=\textwidth]{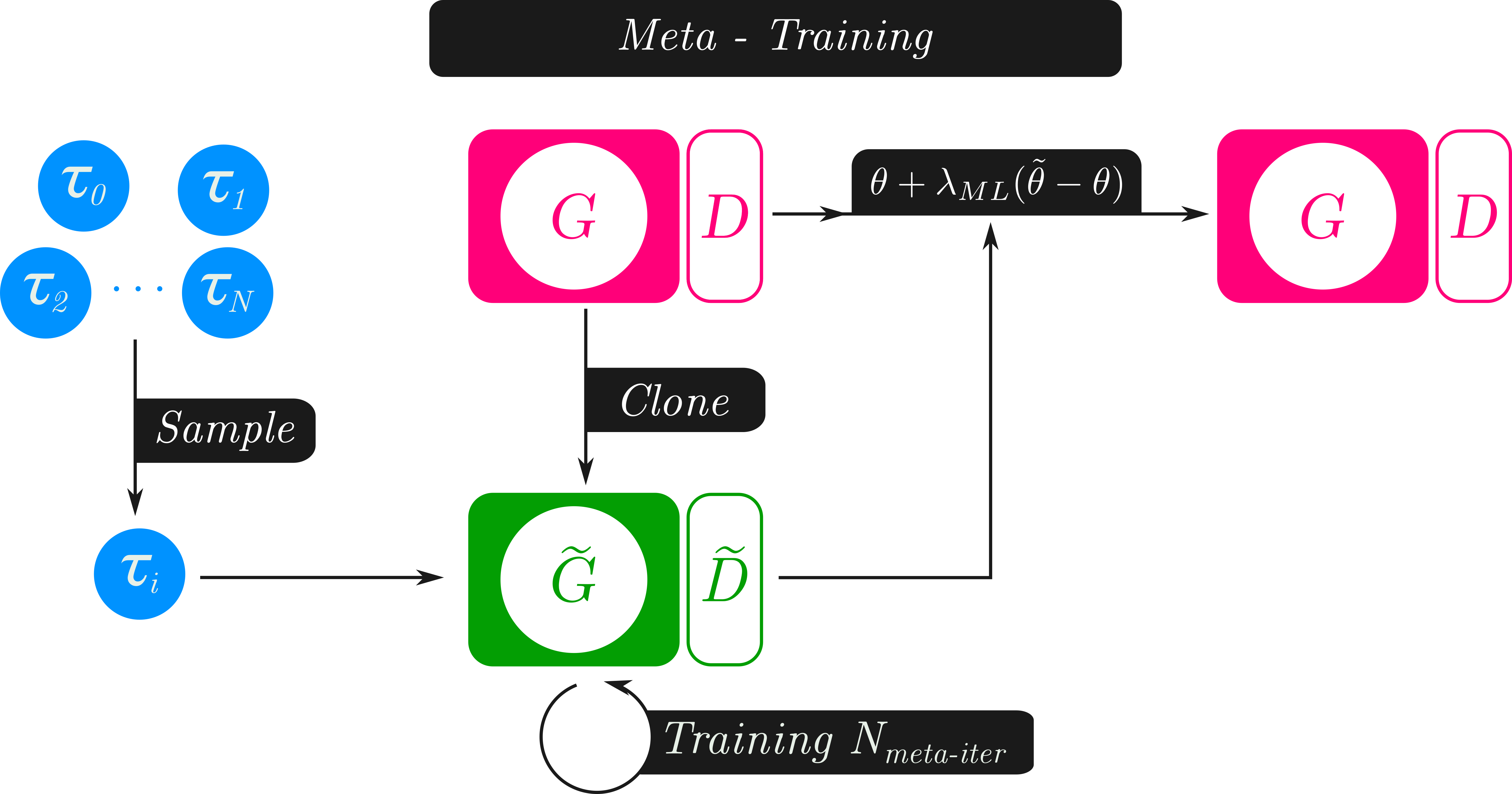}
	\caption{Scheme of a single training epoch. A task $\tau_i$ is sampled and $N_{\mathrm{meta\_iter}}$ inner iterations are performed on cloned networks. Then $\theta$ and $\widetilde{\theta}$ are used to update the networks parameters using the Reptile equation.}
	\label{fig:train}
\end{figure}

The problem setting is as follows. A large dataset of images, called $\mathcal{D}$, 
is used to extract random input images.
Let $\tau_j$ be a single task, where $j$ ranges over the number of chosen training domains, here called $N_\tau$.
Each task dataset consists of a restriction of $\mathcal{D}$ on the images of a single domain, called $\mathcal{D} |_{\tau_j}$. 
Hyper-parameters of the algorithm are the inner learning rates of $G$ and $D$ networks, respectively $\lambda_G$ and $\lambda_D$; the loss weights $w_{\mathrm{adv}}$, $w_{\mathrm{dom}}$, $w_{\mathrm{rec}}$, and $w_{\mathrm{feat}}$; two thresholds $t$ and $T$ for keeping discriminator accuracy into a certain range (in order to neither over- nor under-train $D$); and a learning rate for the outer networks, i.e. a meta-learning rate $\lambda_{ML}$. 
Parameters of the networks, i.e. networks weights and biases, are indicated as $\theta_G$ and $\theta_D$ for $G$ and $D$, respectively.
The algorithm is divided into two phases, as illustrated in Figure \ref{fig:meta_train}. The first one is the training phase, and the second one is the inference phase. In the training phase, the $G$-$D$ network is trained repeatedly on a single task, randomly extracted at each epoch from the set of available tasks.
During inference phase, instead, a new task ${\tau}_I$ is used for a last-time few-shot training to adapt the network to the new domain.
A detailed explanation of training phase is given in the next section and in Figure \ref{fig:train}, and it is followed by another section devoted to the description of the inference phase.


\paragraph{\bf Training}

\begin{algorithm}[h]
	\small
	\setstretch{1.2} 
	\caption{MetalGAN algorithm}\label{alg:metalgan}
	\begin{algorithmic}[1]
		\Require $N_{\mathrm{epochs}}:$ number of epochs
		\Require $N_{\tau}:$ number of selected domains
		\Require $\lambda_{\mathrm{ML}}:$ meta-learning rate
		\State load entire dataset : $\mathcal{D}$
		\State load datasets restricted to each single task $\tau_j$ : $\mathcal{D} |_{\tau_j}$ for $j \in \{0, \dots, N_\tau\}$
		\For{$epoch \in \{ 0, \dots, N_{\mathrm{epochs}} \} $}
		\State extract randomly $\tau_j$			
		\State clone $D$ into $\widetilde{D}$ of parameters $\theta_{\widetilde{D}}$ \label{lin:clone_d}
		\State clone $G$ into $\widetilde{G}$ of parameters $\theta_{\widetilde{G}}$ \label{lin:clone_g}
		\State {\bf Inner training loop} on $\tau_j$
		
		\State $\theta_G \gets \theta_G + \lambda_{\mathrm{ML}} \left( \theta_{\widetilde{G}} - \theta_G \right)$
		\Comment{updates generator parameters}\label{lin:upd_g}
		
		\State $\theta_D \gets \theta_D + \lambda_{\mathrm{ML}} \left( \theta_{\widetilde{D}} - \theta_D \right)$
		\Comment{updates discriminator parameters}\label{lin:upd_d}
		
		\EndFor
	\end{algorithmic}
	\label{metal_alg}
\end{algorithm}

The algorithm for training consists of an outer and an inner loop, similar to Reptile.
The outer loop is responsible of training the actual $G$-$D$ networks, updating their parameters, epoch-by-epoch as a traditional learning algorithm.
At each epoch, a task $\tau$ is randomly sampled from a distribution of tasks. It is recalled that, in our case, a task is a domain and the associated dataset is the few-shot subset of the set of images in the domain.
Then, $G$ and $D$ networks are cloned into $\widetilde{G}$ and $\widetilde{D}$ networks of parameters $\theta_{\widetilde{G}}$ and $\theta_{\widetilde{D}}$, respectively.

The cloned networks are trained in the inner training loop, where the traditional DCGAN training is performed by using task images, here indicated as $y$. This is needed in order to learn the current domain. Also a small portion of generic images from $\mathcal{D}$ is used, to teach the generator to perform domain switch. These images are called $x$.
It is required to the generator to learn the transformation from the random image $x$ of the dataset into an output image ``similar'' to $y$, i.e., the generated image should belong to the extracted task.

Finally, the obtained parameters are used to update $G$ and $D$ weights, with the Reptile rule (a sort of SGD step where the gradient is approximated by the difference between inner and outer weights).
This baseline is illustrated in Figure~\ref{fig:train}.

In detail, the outer training loop is shown in Algorithm~\ref{alg:metalgan}. 
Lines~\ref{lin:upd_g}--\ref{lin:upd_d} of Algorithm~\ref{alg:metalgan} are responsible of the parameter adaptation for networks $G$ and $D$ and such an operation is performed layer by layer.

\begin{algorithm}
	\small
	\setstretch{1.2}
	\caption{Inner training loop}\label{alg:inner_loop}
	\begin{algorithmic}[1]
		\Require $\tau:$ extracted task in the outer loop
		\Require $N_{\mathrm{meta\_iter}}:$ number of inner epochs
		\Require $\lambda_D, \lambda_G:$ learning rates of $D$ and $G$
		\Require $w_{\mathrm{adv}}, w_{\mathrm{dom}}, w_{\mathrm{rec}}, w_{\mathrm{feat}}:$ adversarial, domain, reconstruction, and feature weights
		\Require $t, T:$ minimum and maximum thresholds for discriminator accuracy
		\For {$i \in \{0, \dots, N_{\mathrm{meta\_iter}} \}$}
		\State sample $y$ from $\mathcal{D} |_\tau$
		\State sample $x$ from $\mathcal{D}$
		\State \(\triangleright\) {\bf Discriminator training:}
		\State $
		\varepsilon_{D} \gets \nabla_{\theta_{\widetilde{D}}} \mathcal{L}_{\mathrm{adv}} ( \widetilde{D}, \widetilde{G})
		$ \Comment $x$ is considered fake, $y$ real
		\State $
		\varepsilon_{\mathrm{dom}} \gets \nabla_{\theta_{\widetilde{D}}} 
		\mathcal{L}'_{\mathrm{dom}} (\widetilde{D})
		$ \Comment $x$ is considered false, $y$ true
		\State calculate accuracy $a_{\widetilde{D}}$ of discriminator $\widetilde{D}$
		\If {$ a_{\widetilde{D}} < T$}
		\State $
		\theta_{\widetilde{D}} \gets \theta_{\widetilde{D}} - \lambda_D (\varepsilon_{D} + \varepsilon_{\mathrm{dom}})
		$
		\EndIf
		\If {$ a_{\widetilde{D}} > t $ {\bf or} $i = 0 $}
		\State \(\triangleright\) {\bf Generator training:}
		\State $
		\varepsilon_G \gets 
		\nabla_{\theta_{\widetilde{G}}} 
		\mathbb{E}_{\widetilde{G} ( x ) \sim p_{\tau}} [ \log{ \widetilde{D} ( \widetilde{G} ( x ) ) } ]
		$
		\Comment $\widetilde{G} ( x )$  is considered real
		\State $
		\varepsilon_{\mathrm{task\_rec}} \gets
		\nabla_{\theta_{\widetilde{G}}}
		\mathcal{L}_{\mathrm{rec}} ( \widetilde{G} )
		$ \Comment the reconstruction is made with $y$
		\State $
		\varepsilon_{\mathrm{dom}} \gets \nabla_{\theta_{\widetilde{G}}}
		\mathcal{L}''_{\mathrm{dom}} (\widetilde{D}, \widetilde{G})
		$ \Comment both $y$ and $\widetilde{G}(x)$ are considered true
		\State $
		\varepsilon_{\mathrm{feat}} \gets \nabla_{\theta_{\widetilde{G}}} \mathcal{L}_{\mathrm{feat}}(\widetilde{D}_{\mathrm{feat}}, \widetilde{G})
		$
		\State $
		\theta_{\widetilde{G}} \gets \theta_{\widetilde{G}} - \lambda_G (
			w_{\mathrm{adv}} \varepsilon_G
			+ w_\mathrm{rec} \varepsilon_{\mathrm{rec}} 
			+ w_{\mathrm{dom}} \varepsilon_{\mathrm{dom}} 
			+ w_{\mathrm{feat}} \varepsilon_{\mathrm{feat}})
		$
		\EndIf
		\EndFor
	\end{algorithmic}
\end{algorithm}

The inner training loop is illustrated in Algorithm~\ref{alg:inner_loop}.
It is nothing more than a classic DCGAN training, but performed on the cloned networks. 
For each iteration, a small part of two datasets, that is the task dataset and the full training dataset, is used. The whole $\mathcal{D}$ is sampled randomly only for $N_{\mathrm{meta\_iter}}$ iterations, using only few batches of images.
The chosen task dataset $\mathcal{D} |_{\tau}$ is used for extracting domain specific images.
The first part is the discriminator training. Domain loss and adversarial loss are computed as in Section~\ref{sec:architecture}, and $\widetilde{D}$ parameters are updated if the accuracy of the discriminator is under a certain threshold $T$.
On the contrary, the second part, that is the generator training, is executed only if the accuracy of the discriminator is above a certain threshold $t$. During this step, adversarial, task reconstruction, domain, and feature losses are all employed to update $\widetilde{G}$ parameters.

\paragraph{\bf Inference}

The inference part is also a crucial one. In our work, we experiment the use of few images for adapting the trained model to new, unseen, domains, directly during the inference phase.
The idea is to feed the trained $G$-$D$ networks with images from new domains, moving the obtained parameters $\theta_G$ and $\theta_D$ in a new optimal direction to include the new tasks. A sort of fine-tuning is performed, by showing to the model few images from a new domain, and then few images from another new domain, and so on.
\begin{algorithm}[h]
	\small
	\caption{Inference}\label{alg:inference}
	\begin{algorithmic}[1]
		\Require $\lambda_{\mathrm{ML}}:$ meta-learning rate for inference
		\Require $\mathcal{T} :$ set of {\bf new} tasks/domains
		\Require $N_{\mathrm{inf\_epochs}}, N_{\mathrm{inf\_train}}, N_{\mathrm{inf\_test}} :$ number of inference iterations
		\State load few-shot test dataset: $\mathcal{D}^{(\mathrm{test})}$
		\State load few-shot restricted test dataset on each $\tau \in \mathcal{T} : \mathcal{D}|_{\tau}$
		\State  \(\triangleright\) {\bf Fine-tuning on unseen domains:}
		\For {$epoch \in \{0, \dots, N_{\mathrm{inf\_epochs}} \}$}
		\For {$\tau \in \mathcal{T}$}
		\State clone $G$-$D$ networks
		\State {\bf Inner training loop} on $\tau$, for $N_{\mathrm{inf\_train}}$ iterations
		\State update $\theta_G$ and $\theta_D$ with Reptile rule
		\EndFor
		\EndFor
		\State  \(\triangleright\) {\bf Inference on unseen domains:}
		\For {$\tau \in \mathcal{T}$}
		\State clone $G$-$D$ networks
		\State {\bf Inner training loop} on $\tau$, for $N_{\mathrm{inf\_test}}$ iterations
		\For {$x \in \mathcal{D}^{(\mathrm{test})}$}
			\State generate output image $\widetilde{G}(x)$
		\EndFor
		\State \Comment $G$-$D$ parameters are not updated anymore
		\EndFor
	\end{algorithmic}
	\label{alg:rec}
\end{algorithm}

The settings of the inference algorithm are the following. 
A set of unseen tasks (or domains) $\mathcal{T}$ is adopted.
A test dataset $\mathcal{D}^{(\mathrm{test})}$ containing all domains, where $\mathcal{D}^{(\mathrm{test})} \cap \mathcal{D} = \emptyset $, is used.
As for the training dataset, for each new domain to infer $\tau \in \mathcal{T}$, the adequate restriction of dataset is used, $\mathcal{D}|_{\tau} \subset \mathcal{D}$, in order to avoid overlaps between test and training datasets.

Algorithm~\ref{alg:inference} shows the inference method.
It is divided in two main parts: a few-shot fine-tuning, and a test phase where unseen images are transformed into target domain images.
In the first part, new domains are used to learn a new parameter adaptation, using few images per class. Moreover, the meta-learning rate for inference is greater than the one used in training, in order to ensure a faster adaptation.
The second part is used only for generating the results, and it resembles a more classic inference. The inner training loop of the meta-learning algorithm is used, but obtained parameters are not updated for the next domain.

\begin{figure}[t]
	\centering
	\includegraphics[width=\textwidth]{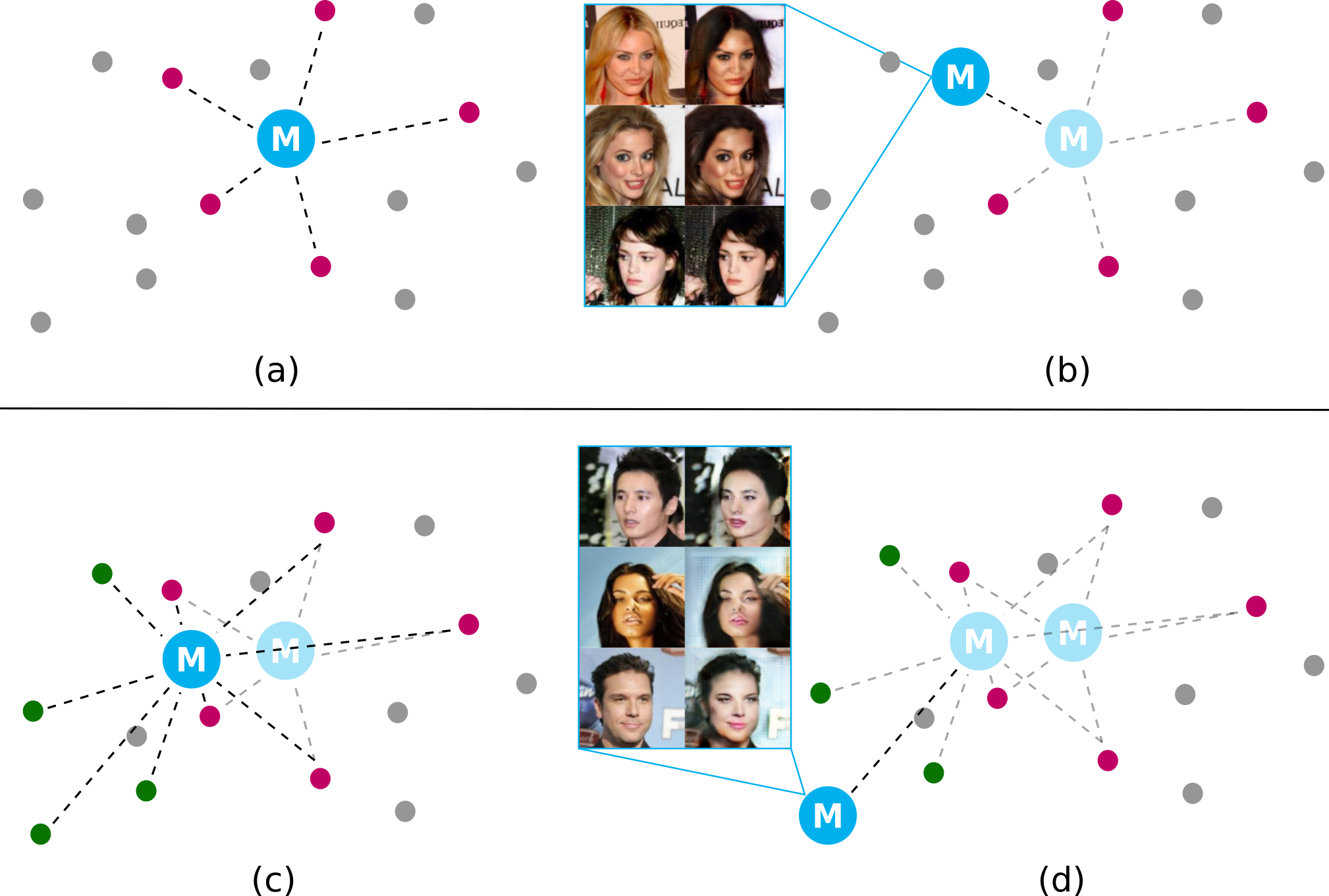}
	\caption{Inference algorithm in the case of one of the training domains, `Black Hair' (b) and in the case of an unseen domain, `Heavy Makeup' (d). The violet dots represent the domains used during training while the green dots are the unseen domains. The model $M$ can move towards a known model, i.e. from state (a) to state (b); the other option is fine-tuning towards a set of unseen domains (c), then converge to a specific one (d).}
	\label{fig:inference}
\end{figure}
An explanation of inference algorithm is visually provided in Figure~\ref{fig:inference} where two exemplar cases are shown: the top row of the figure (Figure \ref{fig:inference}(a) and \ref{fig:inference}(b)) shows the case of the seen domain `Black Hair', whereas the bottom row (Figure \ref{fig:inference}(c) and \ref{fig:inference}(d)) shows the case of unseen domain `Heavy Makeup'. 
In the image, the model, called $M$, consists of the $G$-$D$ networks trained as in Algorithm~\ref{alg:metalgan} on five different example domains. 
In Figure \ref{fig:inference}(a), a n\"{a}ive illustration of the domain space is provided. Given the dots as domains, and the violet dots as the seen domains, the model $M$ has learned, during training, a sub-optimal representation for the seen domains.
Intuitively, such a representation permits the model to easily move towards the optimal one in few steps of the inner training loop. 
In Figure \ref{fig:inference}(b), an example inference on `Black Hair' domain is shown.
Since `Black Hair' belongs to the trained domains set, the model is cloned and with some steps of inner training loop, the cloned model learns an optimal representation for the domain.
Finally, parameters of the cloned model are not saved nor used for updating the main model, so the situation returns to the one depicted in Figure \ref{fig:inference}(a), ready to make another inference.
When a domain, or a set of domains, are completely new to the generator and discriminator, a preliminary fine-tuning is needed. In Figure \ref{fig:inference}(c), green dots represents these unseen domains. The inference pre-training moves the (already trained) model in a sub-optimal position for both the trained domains and the new domains. In Figure \ref{fig:inference}(d), the inference on the updated model is shown. As Figure \ref{fig:inference}(d) shows, the fine-tuning for the unseen domains moves the model M in a new ``position'' in the space, closer to the unseen domains. In this way, during the inference (Figure \ref{fig:inference}(d)) the `Heavy Makeup' domain is better learnt and more correct images are generated, even if the domain has not been seen during the training.

Please note that the figure is completely exemplifying, since it depicts domains in random positions, and does not take in account the intersection between them, nor their `real' positioning in an actual domain space (which is unknown). The idea of the model moving towards a sub-optimal yet effective representation during training epochs---minimizing the expected distance from all tasks---, and an informal proof of the idea using Euclidean distances in the manifold of optimal solutions of a task, is provided in Reptile paper~\cite{nichol2018first}.

\section{Experimental Results}\label{sec:experiments}

This section presents visual and quantitative results of performed experiments of MetalGAN, compared with StarGAN ones.
All our experiments were conducted using the CelebA dataset \cite{liu2015faceattributes} which is a large-scale face attributes dataset with more than 200k celebrity images, each with 40 attribute annotations.
We decided to test our algorithm on this dataset for three main reasons: first of all, since it contains images of faces with all kind of attributes, it is suitable for multi-domain image-to-image task; secondly, it was used by StarGAN so it allows a clear comparison between the results of the two different algorithms; and finally, even though our approach is completely label-less, it is very easy to automatically divide a-priori the dataset in its different domains.

Experiments are divided into two categories: test results on seen domains (i.e., tasks the $G$-$D$ networks were trained on), and results on unseen domains.
In case of experiments on StarGAN, since their algorithm requires labels, we trained their network on some domains with a few number of images (1000), and we call these ``unseen'' domains. This workaround permits us to compare StarGAN with our inference on unseen domains. It is worth to note that this approach is unfair for us, since StarGAN is fully trained for each of these ``unseen'' domains, while we only perform a small inference step. This is due to the fact that we can choose to add new domains to our network at every time, while StarGAN needs to define all the domains at the training stage.

\subsection{Results on Trained Domains}
\label{sec:results_trained}

\begin{table}[h]
	\caption{Hyper-parameters of MetalGAN training phase.}
	\begin{center}
		\begin{tabular}{|l|c|}
			\hline
			$N_{\mathrm{epochs}}$ & 100000 \\
			\hline
			$\lambda_{\mathrm{ML}}$ &  0.01 \\
			\hline
			$\lambda_{G}$, $\lambda_{D}$ (Adam) & 0.0001 \\
			\hline
			$N_{\mathrm{meta\_iter}}$ & 20 \\
			\hline
			batch size & 16 \\
			\hline
			$w_{\mathrm{adv}}$ & 1 \\
			\hline
			$w_{\mathrm{dom}}$ & 1 \\
			\hline
			$w_{\mathrm{rec}}$ & 10 \\
			\hline
			$w_{\mathrm{feat}}$ & 1 \\
			\hline
		\end{tabular}
		\label{tab:exp_settings}
	\end{center}
\end{table}

We trained $G$-$D$ networks model on 5 domains, namely `Eyeglasses', `Male', `Blond Hair', `Black Hair', and `Pale Skin' for $N_{epochs} = 100000$ using the MetalGAN algorithm, and on the same domains for $200000$ epochs using StarGAN.

\begin{figure}[h]
	\begin{subfigure}[t]{.53\linewidth}
		\includegraphics[height=0.44\textheight]{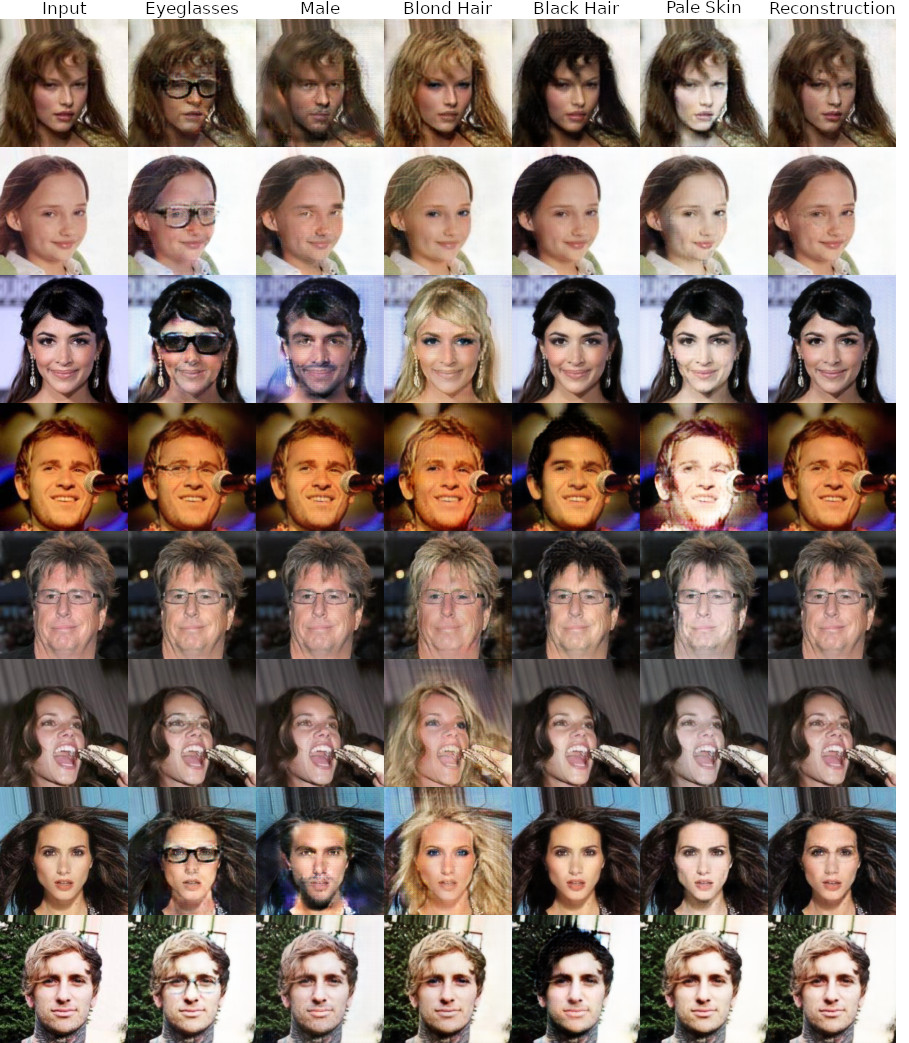}
		\caption{MetalGAN outputs.}
	\end{subfigure}
	\begin{subfigure}[t]{.3\linewidth}
		\includegraphics[height=0.44\textheight]{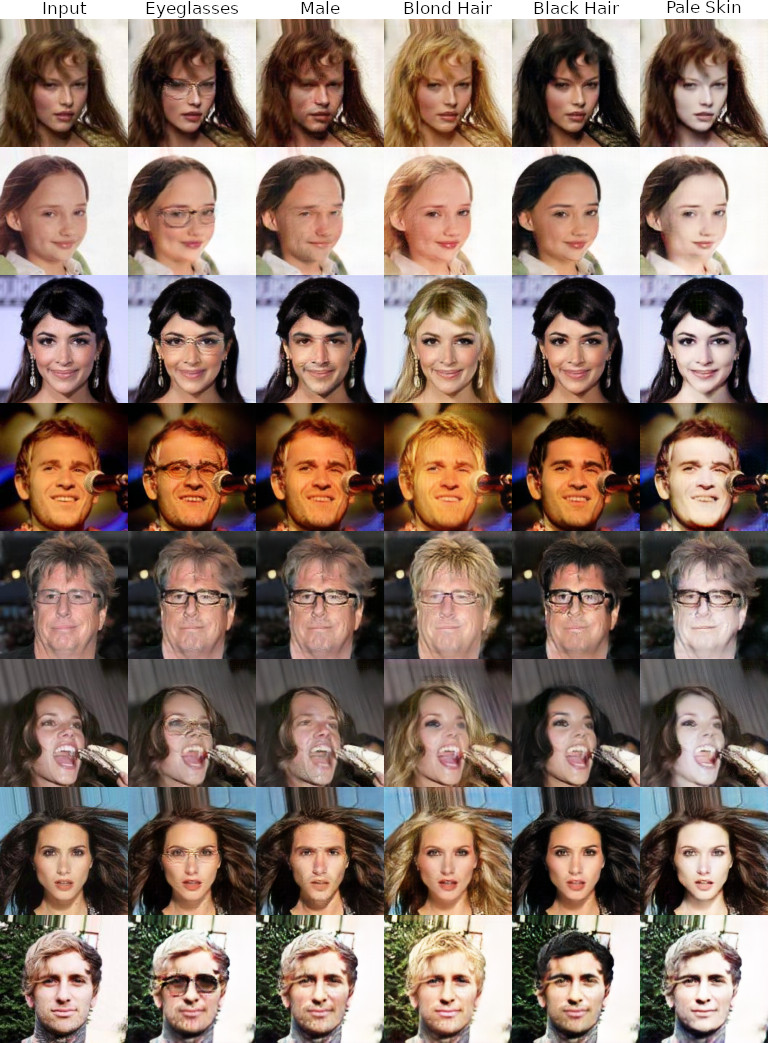}
		\caption{StarGAN outputs.}
	\end{subfigure}
	\caption{Results on training classes. In the first column, the input images. From second to fifth column there are the outputs of the model moved towards the respective domain, in case of MetalGAN, or labeled with the indication of the domain, in case of StarGAN. The last column of MetalGAN results is the output of the model without moving it from the sub-optimum.}
	\label{fig:training_results}
\end{figure}

Table \ref{tab:exp_settings} presents the main settings for our experiments. We set the Reptile learning rate $\lambda_{\mathrm{ML}}$ to 0.01 and optimized the generator and discriminator networks using Adam with a learning rate equals to 0.0001. Furthermore, we set the number of meta-iterations $N_{meta\_iter}$ during training equals to 20 since we empirically found that this value represents the best trade-off between speed and accuracy of the algorithm. 
For coherence with StarGAN, batch size is set to 16 during training.
Weights for MetalGAN objective during training are left to 1 except for reconstruction weight, that is set to 10, in order to obtain an accurate reconstruction of the image and gain more quality in results.

\begin{figure}[t]
	\centering
	\includegraphics[width=\textwidth]{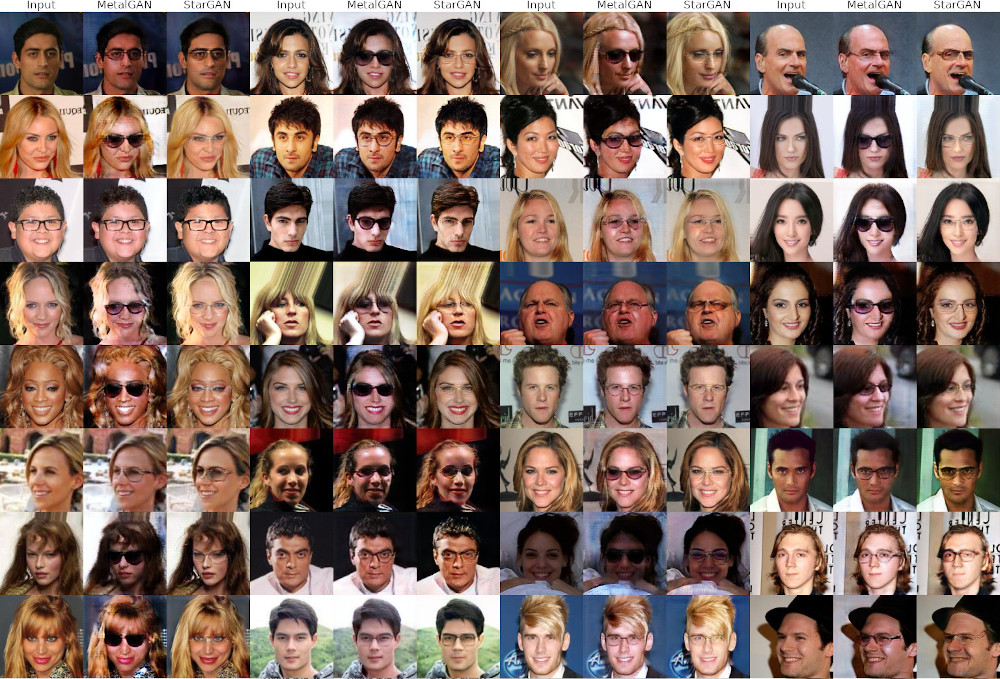}
	\caption{Results on training domain Eyglasses. The image triplets are composed by input image, MetalGAN output and StarGAN output.}
	\label{fig:eyeglasses_inference}
\end{figure}

\begin{figure}[p]
	\centering
	\includegraphics[width=\textwidth]{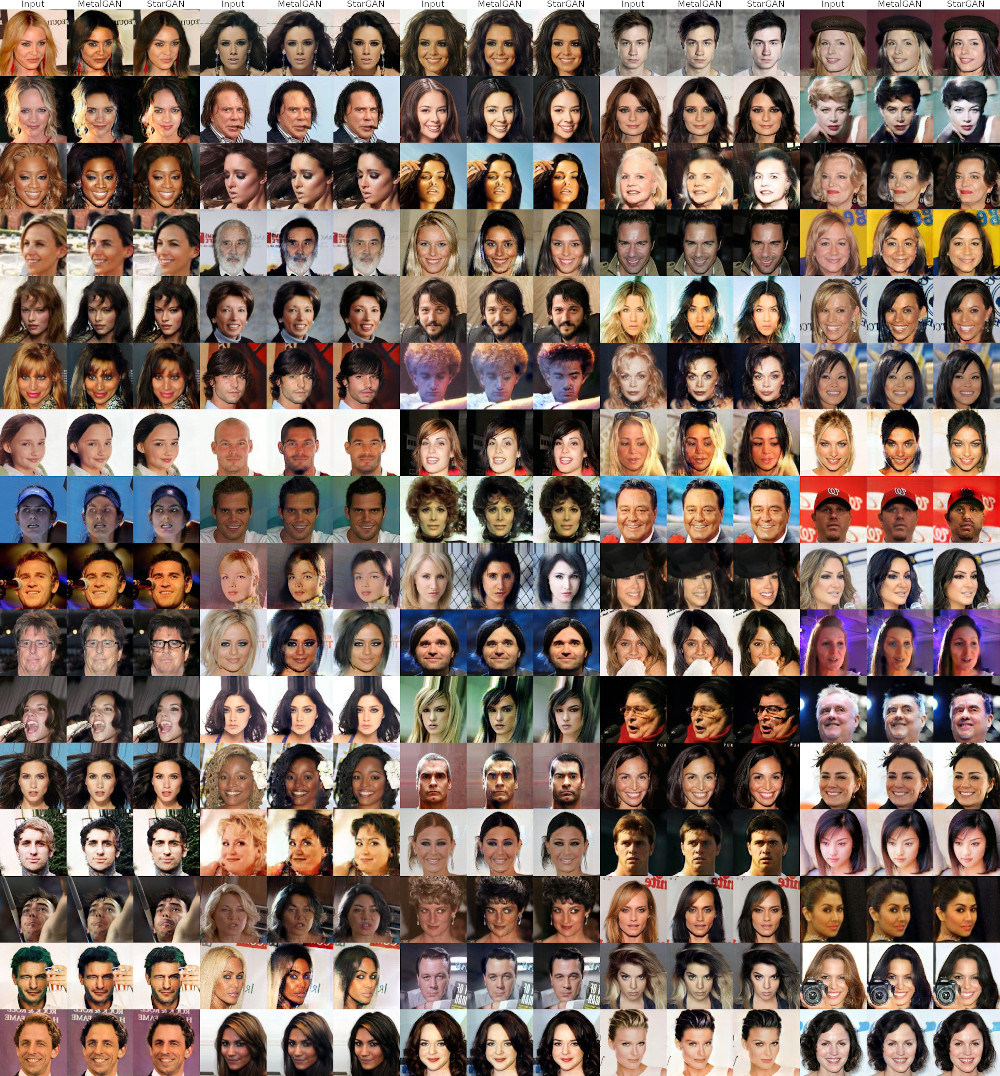}
	\caption{Results on training domain Black Hair. The image triplets are composed by input image, MetalGAN output and StarGAN output.}
	\label{fig:black_hair_inference}
\end{figure}

Figure~\ref{fig:training_results} shows some visual results on a batch of eight input images. Figure~\ref{fig:training_results}(a) contains the outputs of MetalGAN algorithm, while Figure~\ref{fig:training_results}(b) shows the StarGAN outputs. In addition, a greater number of results on some of the training classes are shown in Figure \ref{fig:eyeglasses_inference} and \ref{fig:black_hair_inference}. Figure~\ref{fig:eyeglasses_inference} shows generated images on `Eyeglasses' domain, where input images are put side-by-side to MetalGAN outputs and StarGAN outputs. In the same fashion, results on `Black Hair' domain are reported in Figure~\ref{fig:black_hair_inference}. We decided to choose these two domains since they are very different in terms of features and since our method performs very well on `Eyeglasses' and, on the contrary, it is not so good on `Black Hair'. It is also worth noting that MetalGAN on `Eyeglasses' produces a great variability of examples, compared to StarGAN, generating both simple glasses and sunglasses.

However, the image generation should be considered successful and visually close to StarGAN one.
As a matter of fact, we can see how our label-less approach produces results that are visually very similar to the ones produced by StarGAN. In particular, our algorithm is able to understand the different target domains just by seeing few examples of them each epoch, and can correctly produce these domains from the input images even without labels or supervision. 

\begin{figure}[h]
	\centering
	\includegraphics[width=0.7\textwidth]{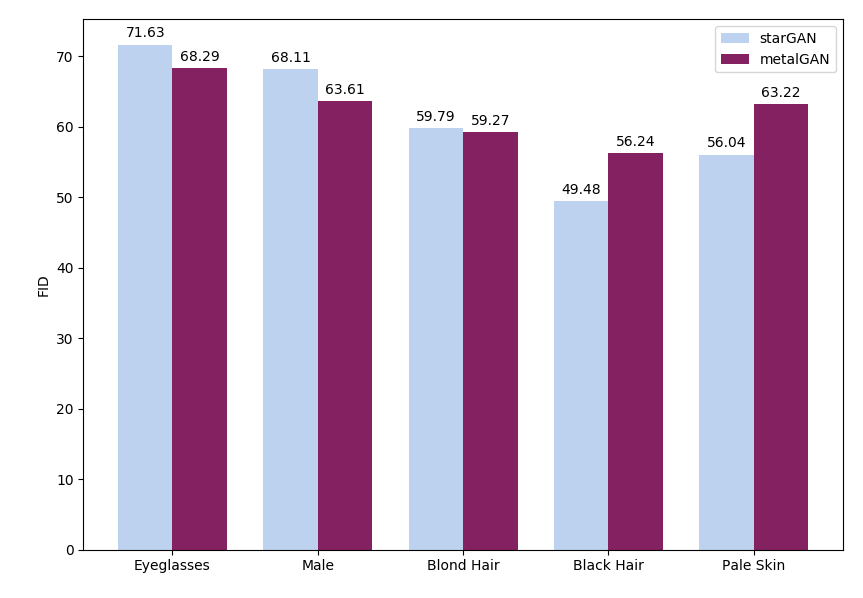}
	\caption{FID score on training domains (the lower the better).}
	\label{fig:Training_graph}
\end{figure}

In addition, we performed quantitative analysis of the produced results. 
As far as we know, no pure theoretical framework is available for a precise quantification of our model contributions and advantages, in order to compare it to others, but there exists some relevant metrics that are suitable for a numerical placement of our proposal. Metrics considered in this work are 
FID (Frechet Inception Distance)~\cite{heusel2017gans} and PRD (Precision and Recall for Distributions)~\cite{sajjadi2018assessing}, described below.
We use FID to calculate the distribution matching between the original CelebA images for each training domains and our results, and we compare our score with the one obtained on StarGAN images. This comparison is presented in Figure \ref{fig:Training_graph} with lower values indicating the better scores. Our method performs slightly better than StarGAN for `Eyeglasses', `Male' and `Blond Hair' domains and slightly worse than StarGAN for `Black Hair' and `Pale Skin', confirming the visual evaluation of the images.

Another quantitative analysis is based on PRD for both StarGAN and MetalGAN methods, using classes of images of CelebA as target datasets. Precision is a measure of raw quality of generated images, and does not take in account the internal variability of the distribution, while recall measures how well the generated images resembles the ``class distribution'' of the target dataset. We choose to measure a single domain at once.
\begin{figure}[h]
\begin{subfigure}{0.2\textwidth}
	\includegraphics[width=\linewidth]{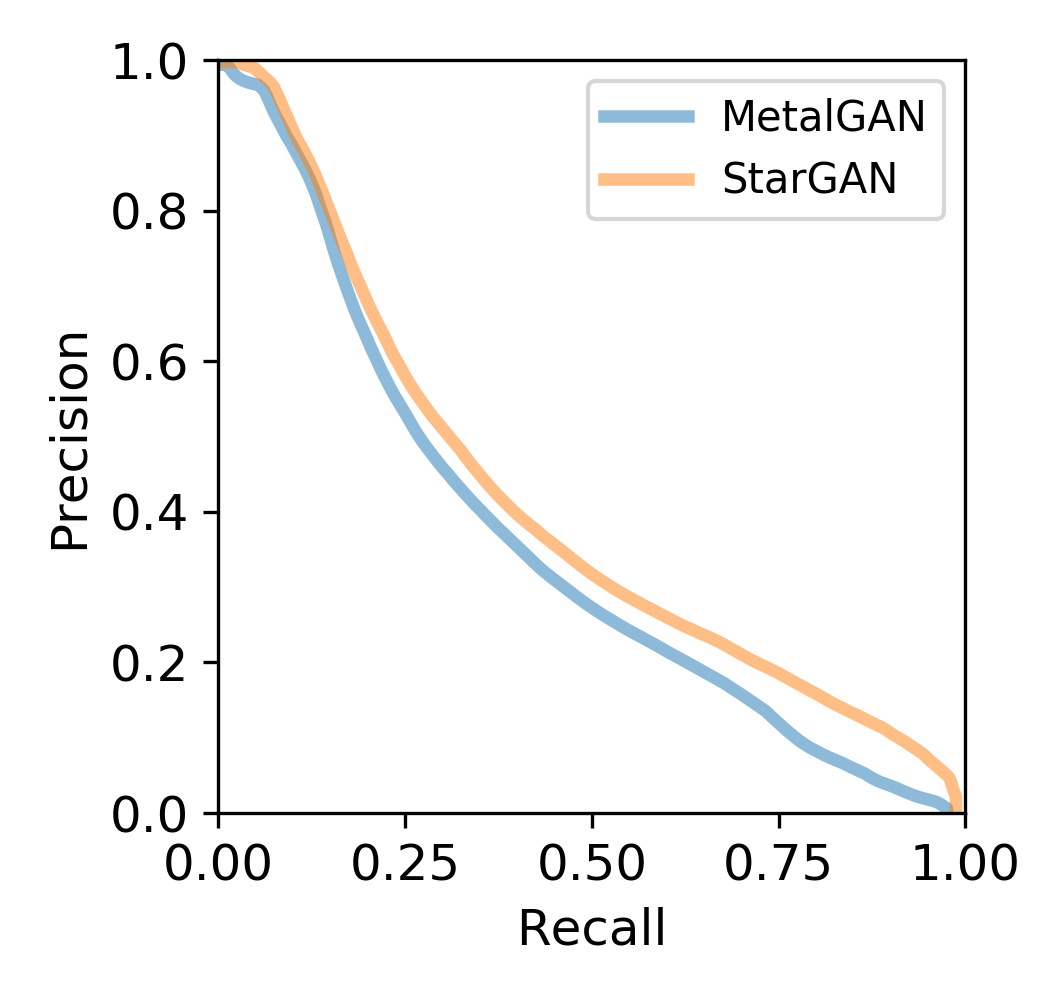}
	\caption{Eyeglasses}
\end{subfigure}%
\begin{subfigure}{0.2\textwidth}
	\includegraphics[width=\linewidth]{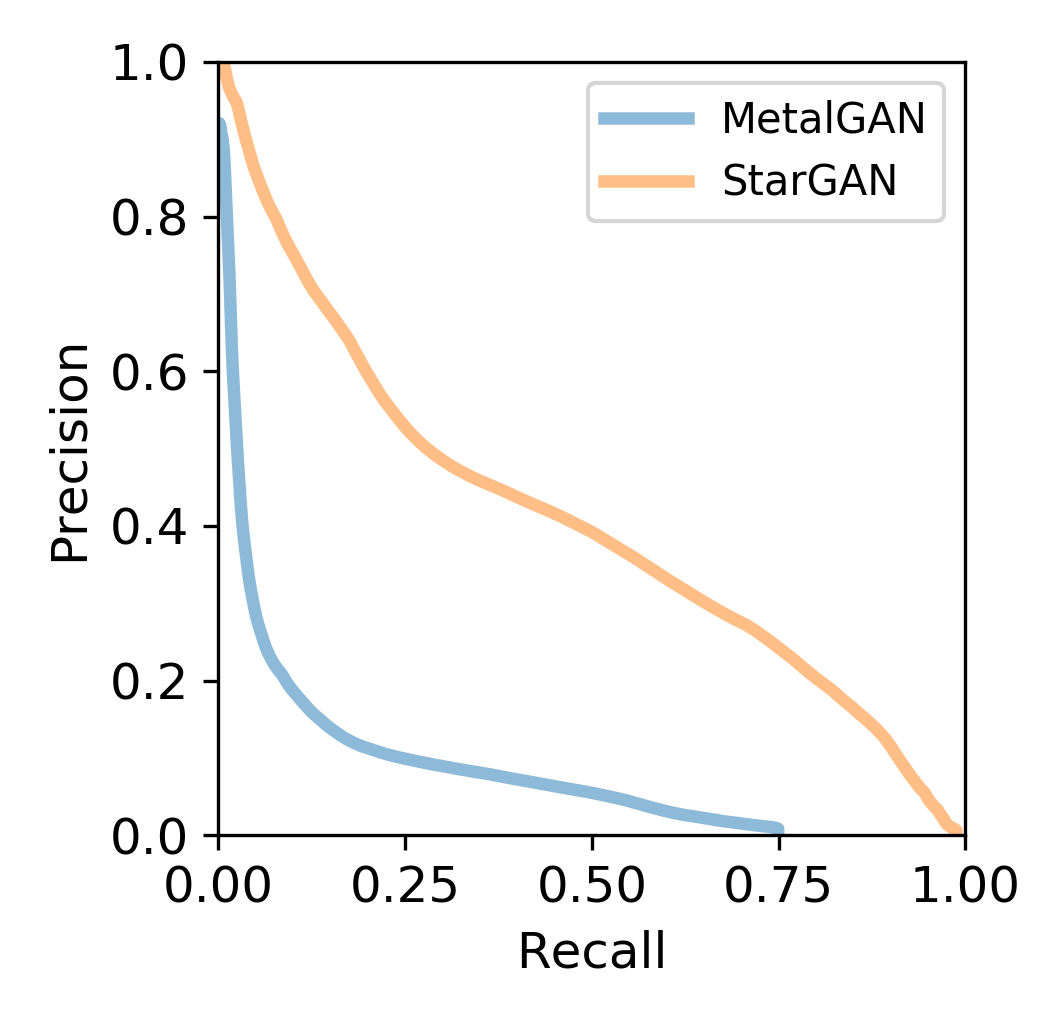}
	\caption{Male}
\end{subfigure}%
\begin{subfigure}{0.2\textwidth}
	\includegraphics[width=\linewidth]{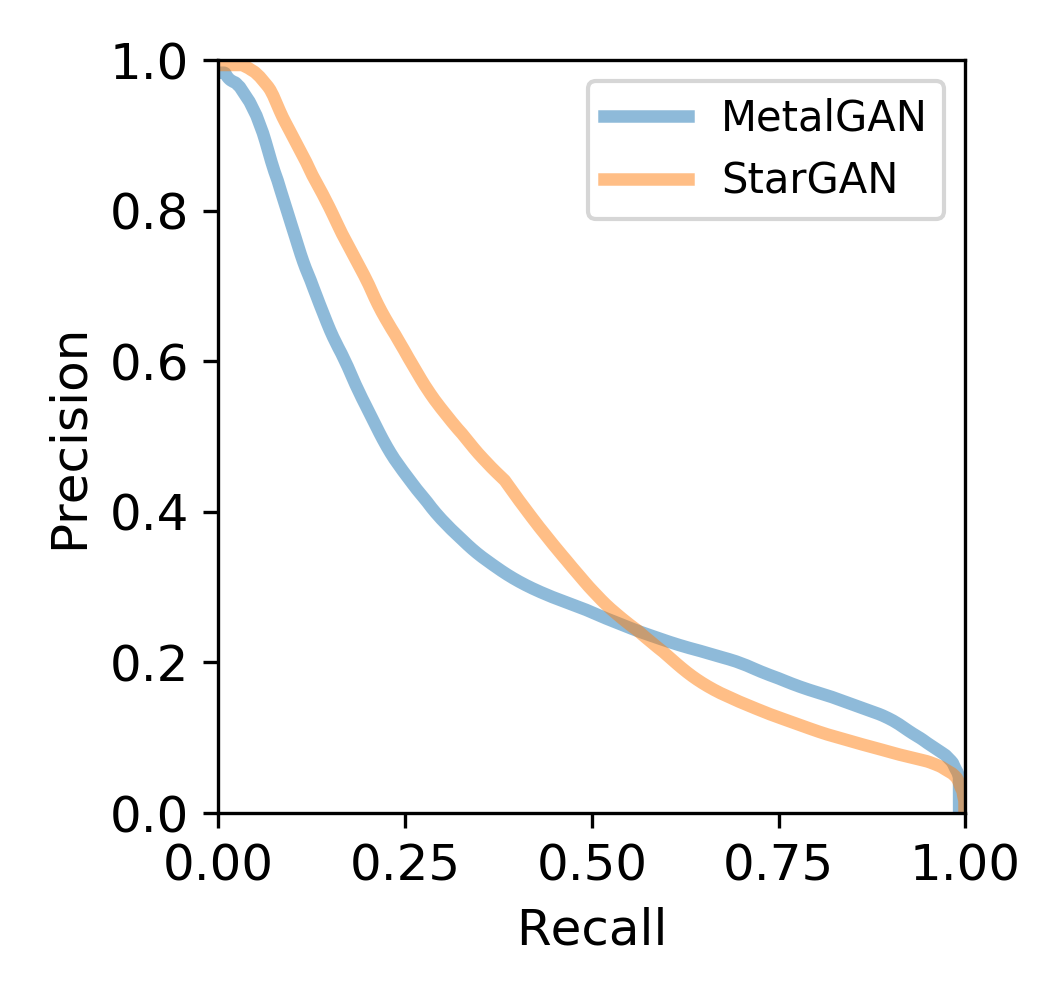}
	\caption{Blond Hair}
\end{subfigure}%
\begin{subfigure}{0.2\textwidth}
	\includegraphics[width=\linewidth]{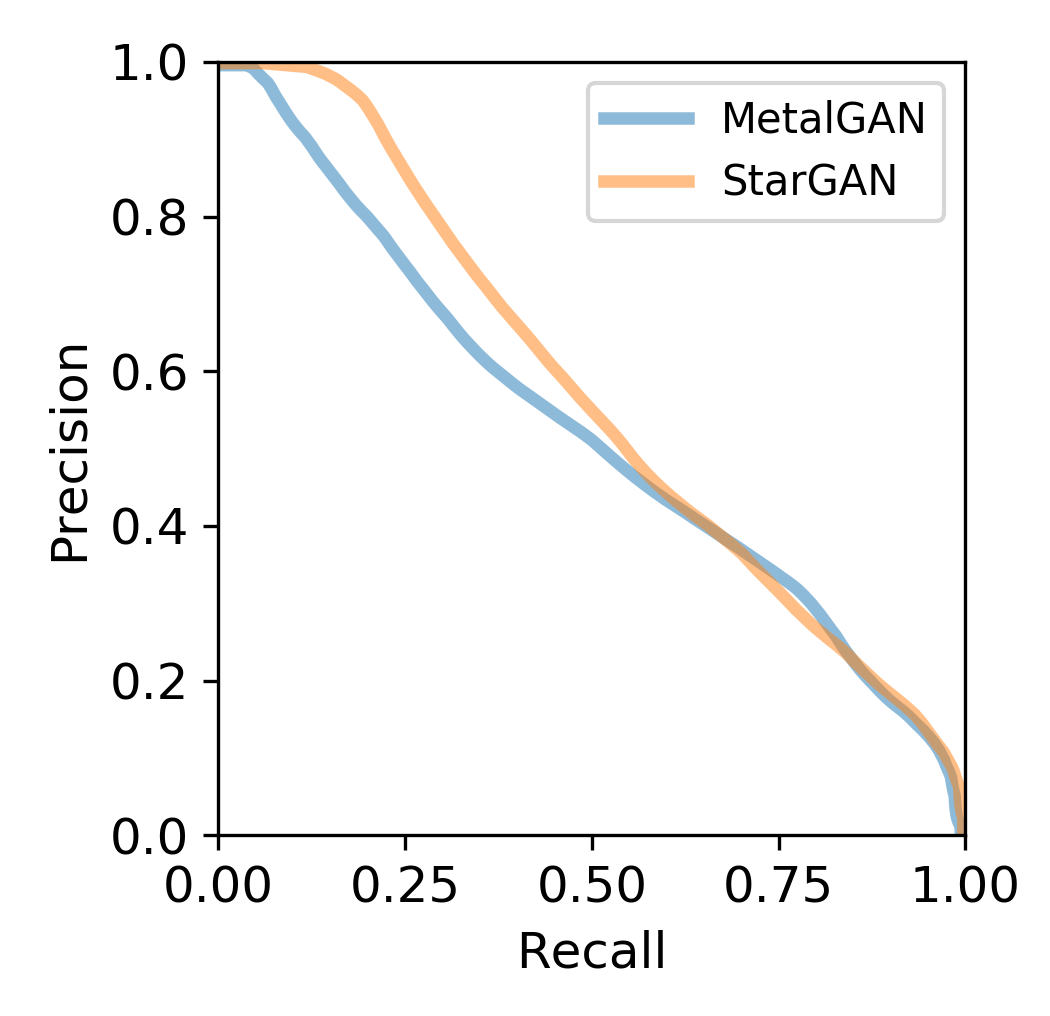}
	\caption{Black Hair}
\end{subfigure}%
\begin{subfigure}{0.2\textwidth}
	\includegraphics[width=\linewidth]{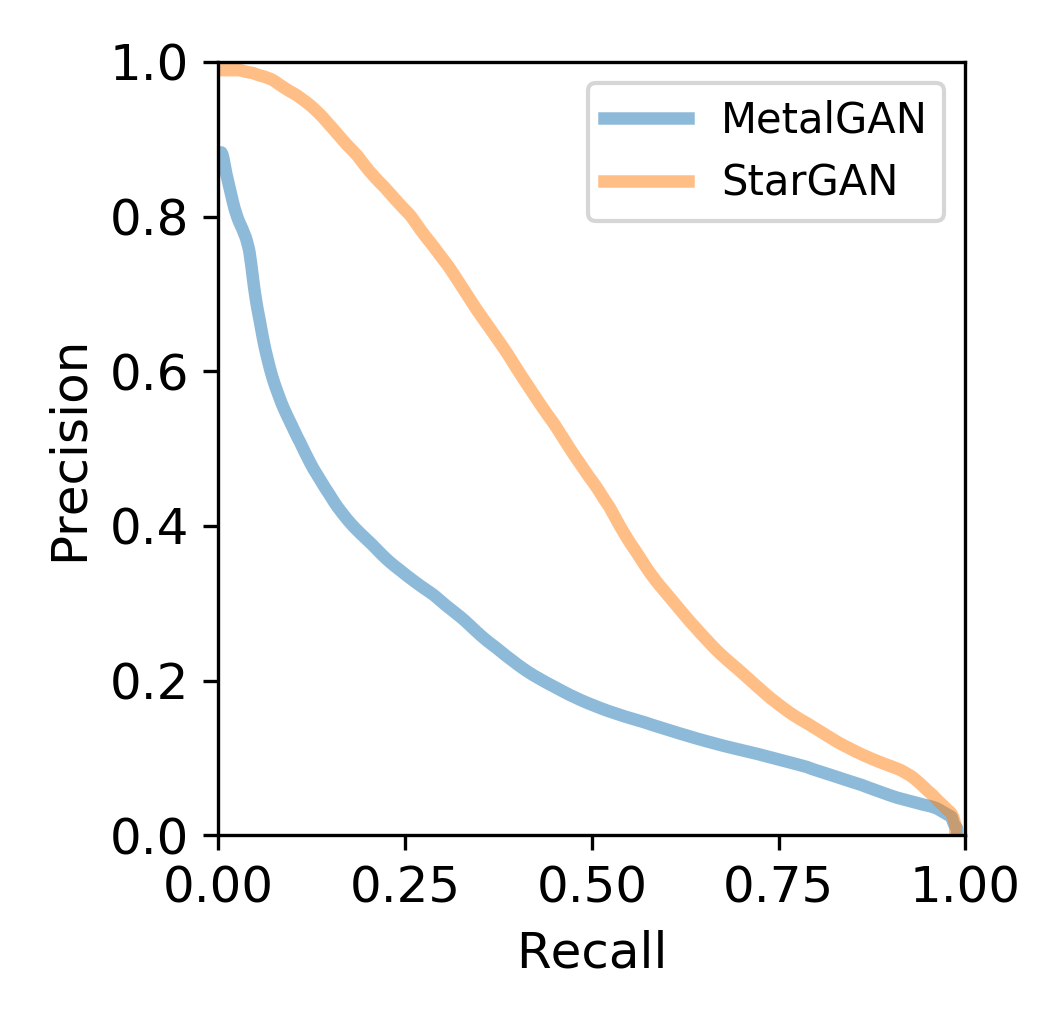}
	\caption{Pale Skin}
\end{subfigure}
\caption{PRD on each training domains.}
\label{fig:prd_training_graph}
\end{figure}
Results are shown in five different graphics, in Figure~\ref{fig:prd_training_graph}.
As shown, MetalGAN PRD on `Eyeglasses', `Blond Hair', and `Black Hair' are very similar to each other and close to StarGAN results. The main difference between StarGAN and MetalGAN in case of hair domains is that StarGAN is usually more precise (it produces images with a better quality w.r.t. the target distribution), but it has a lower recall, meaning that the distribution of StarGAN generated images is less varied than MetalGAN one. Regarding `Male' and `Pale Skin', the precision of MetalGAN suffers from the fact that such domains require a significant change in all the faces in the input image, highlighting a weakness in MetalGAN global reconstruction. On the other hand, the domain change is successful, as confirmed for `Male' FID score.
\begin{figure}[h]
\centering
\includegraphics[width=0.3\textwidth]{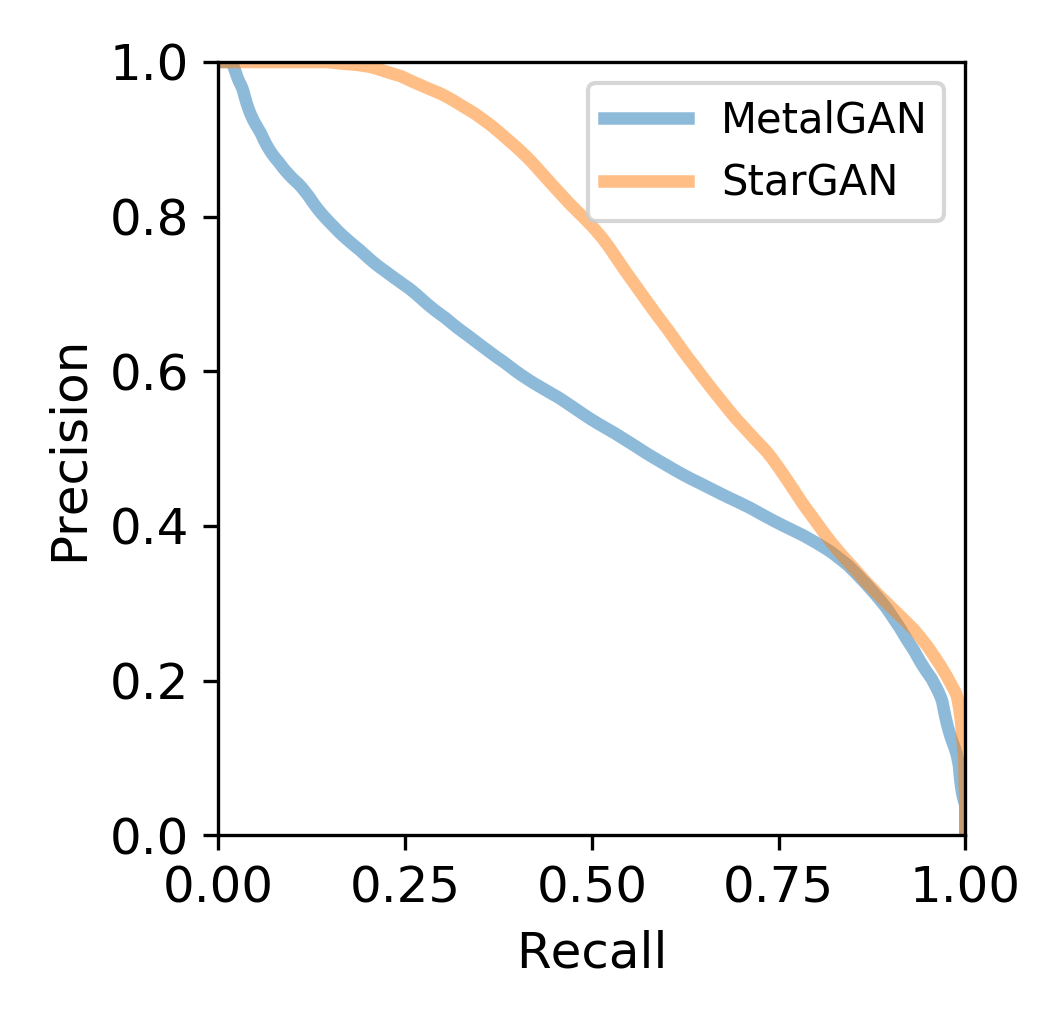}
\caption{Global PRD on training domains.}
\label{fig:prd_alltraining_graph}
\end{figure}
In Figure~\ref{fig:prd_alltraining_graph}, global PRD, computed on all training domains at once, resembles the previous consideration, showing a worse precision of MetalGAN generated distribution, but a similar high recall for StarGAN and MetalGAN distributions.

It is worth emphasizing once more that MetalGAN achieves these results without labels, showing in any case comparable quantitative results and often better qualitative results.

\subsection{Results on Unseen Domains}
\label{sec:result_inference}

\begin{table}[h]
\caption{Hyper-parameters of MetalGAN inference phase.}
\begin{center}
\begin{tabular}{|l|c|}
	\hline
	$N_{\mathrm{epochs}}$ & 10 \\
	\hline
	$\lambda_{\mathrm{ML}}$ & 0.1 \\
	\hline
	$\lambda_{G}$, $\lambda_{D}$ (Adam) & 0.0001 \\
	\hline
	$N_{\mathrm{inf\_train}}$ (train) & 20 \\
	\hline
	$N_{\mathrm{inf\_test}}$ (test) & 100 \\
	\hline
	batch size & 16 \\
	\hline
	$w_{\mathrm{adv}}$ & 100 \\
	\hline
	$w_{\mathrm{dom}}$ & 100 \\
	\hline
	$w_{\mathrm{rec}}$ & 1 \\
	\hline
	$w_{\mathrm{feat}}$ & 1 \\
	\hline
\end{tabular}
\label{tab:inference_exp_settings}
\end{center}
\end{table}

During the inference step, we modify the hyper-parameters of MetalGAN as shown in Table \ref{tab:inference_exp_settings}. In particular, in order to allow the network to quickly adapt to the new domains, we increment the $\lambda_{\mathrm{ML}}$ to 0.1 and we set $w_{\mathrm{adv}}$ and $w_{\mathrm{dom}}$ to 100. Furthermore, since the network already learned to reconstruct the content of the input images we lower $w_{\mathrm{rec}}$ to 1. 

Finally, we tested the MetalGAN trained model on 6 unseen domains, namely `Big Lips', `Bushy Eyebrows', `Heavy Makeup', `Smiling', `Gray Hair', and `Mustache' using the MetalGAN inference.
MetalGAN, trained on the 5 seen domains of Section~\ref{sec:results_trained}, performs 10 further outer iterations (on each new domain), each of them consisting of 20 inner iteration, where $320$ task images are seen for the first time. In this way, a fine-tuned model is obtained, as in Figure~\ref{fig:inference}(c). Then, images are generated by specializing the fine-tuned model on the chosen domain, as in Figure~\ref{fig:inference}(d). Such a specialization is done performing 100 inner iterations per domain.

On the other side, we trained 6 \emph{new} StarGAN models with the same domains used during training, plus one unseen domain for each model, i.e. we obtained a StarGAN model specialized also in `Big Lips', another StarGAN model specialized also on `Bushy Eyebrows', and so on. 
This is necessary, since StarGAN uses image labels, so adding a new domain is possible only by retraining the model. All six new StarGAN models were fully trained for $N_{\mathrm{epochs}} = 200000$.
For StarGAN, ``unseen'' means that only 1000 input images are selected for that domain, as already described in the beginning of Section~\ref{sec:experiments}. 

\begin{figure}[p]
	\centering
	\includegraphics[width=\textwidth]{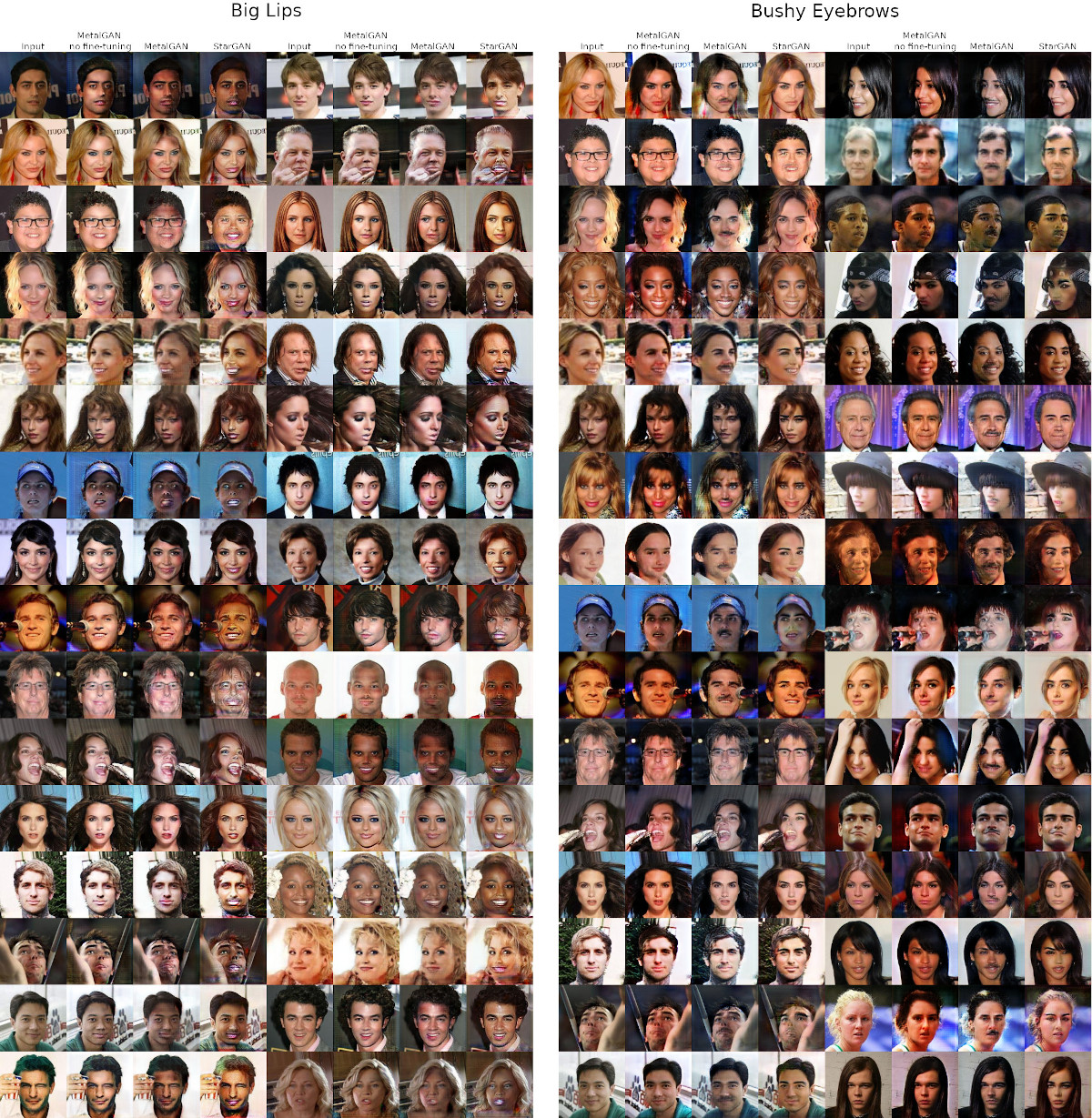}
	\caption{Results on unseen domains Big Lips and Bushy Eyebrows. The image triplets are composed by input image, MetalGAN output without fine-tuning, MetalGAN output with fine-tuning and StarGAN output.}
	\label{fig:biglips_inference}
\end{figure}

\begin{figure}[p]
	\centering
	\includegraphics[width=\textwidth]{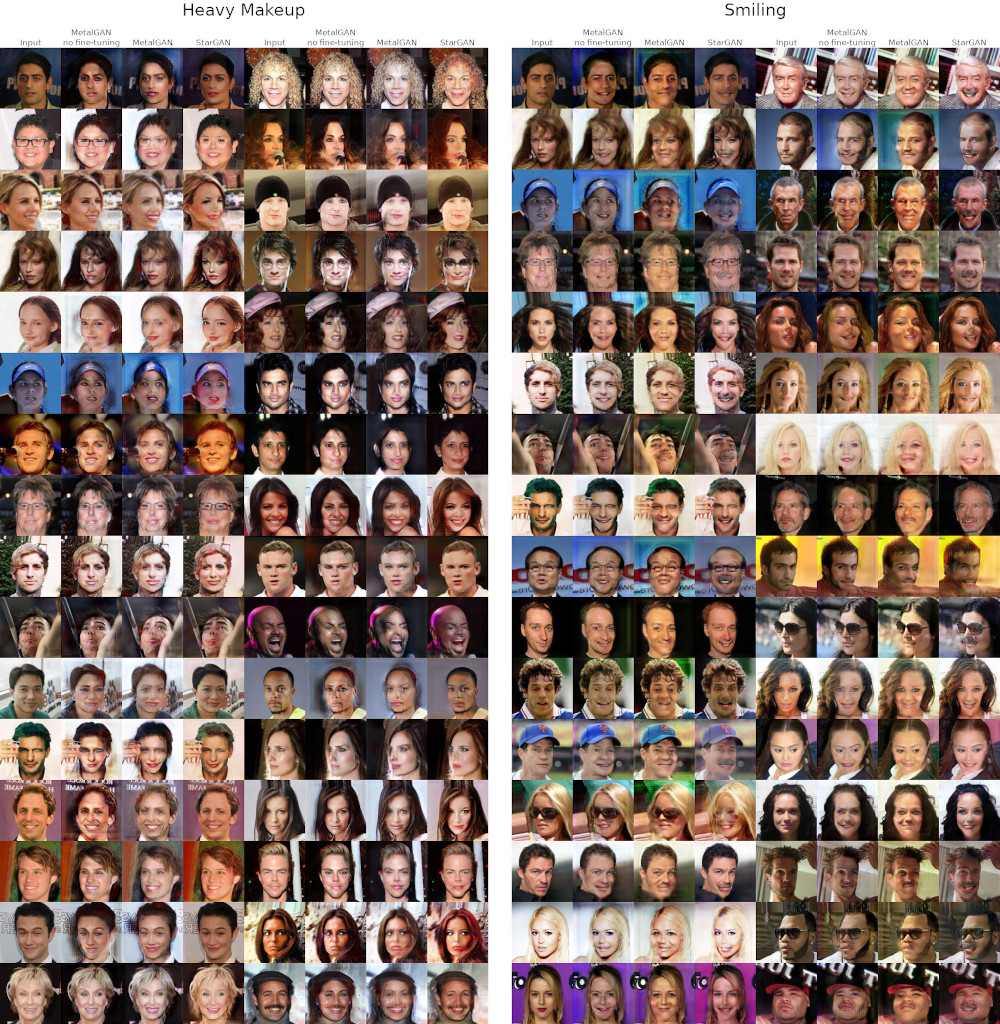}
	\caption{Results on unseen domains Heavy Makeup and Smiling. The image triplets are composed by input image, MetalGAN output without fine-tuning, metalGAN output with fine-tuning and starGAN output.}
	\label{fig:heavymakeup_inference}
\end{figure}

\begin{figure}[p]
	\centering
	\includegraphics[width=\textwidth]{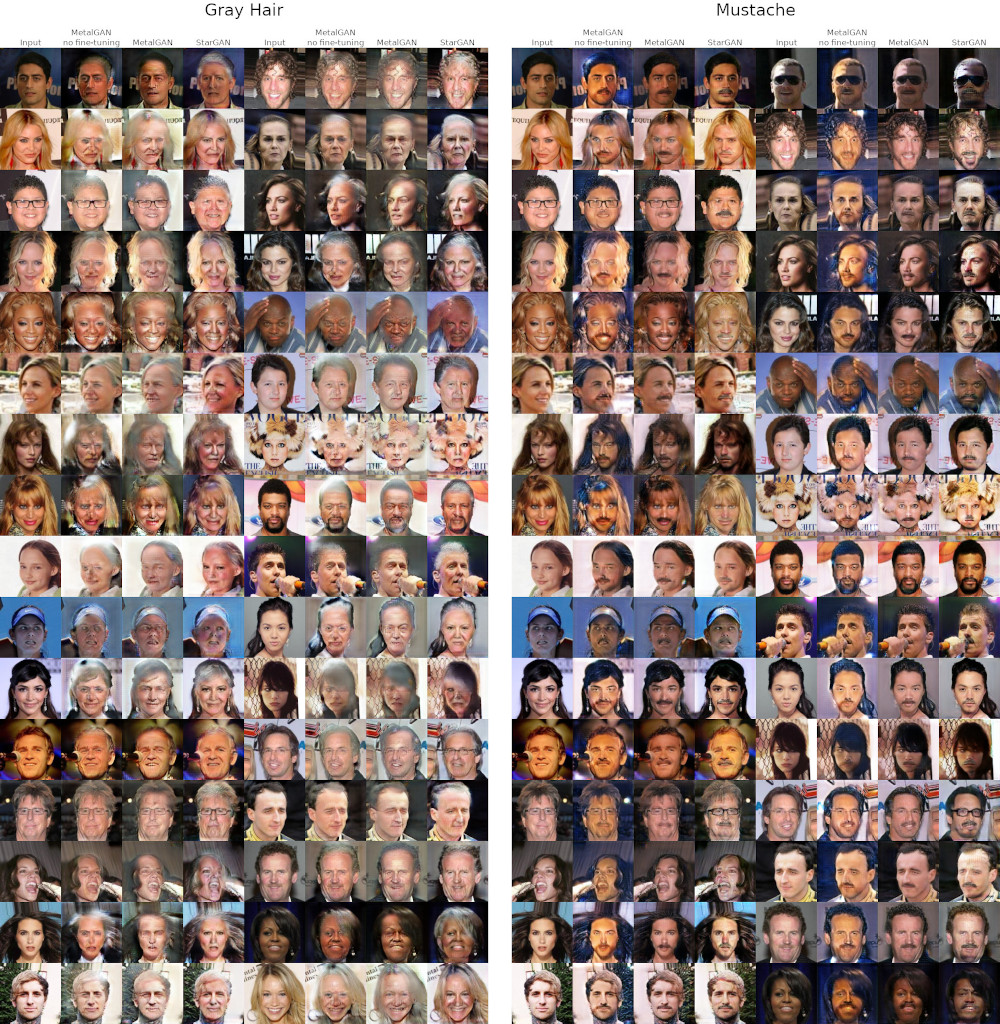}
	\caption{Results on unseen domains Gray Hair and Mustache. The image triplets are composed by input image, MetalGAN output without fine-tuning, metalGAN output with fine-tuning and starGAN output.}
	\label{fig:mustache_inference}
\end{figure}

Visual qualitative results for unseen domains for both MetalGAN and StarGAN are presented in Figure \ref{fig:biglips_inference}, \ref{fig:heavymakeup_inference} and \ref{fig:mustache_inference}.
In particular, for MetalGAN, the results produced without performing the fine-tuning iterations are also shown. Our algorithm is able to produce compelling images even in this case and further improves the visual appearance of the images after the fine-tuning step. In addition, MetalGAN applyes the unseen domains to the input images in a more soft and natural way than StarGAN. This is particularly evident in `Big Lips' and `Smiling', where StarGAN produced results that could be described as ``creepy''.
A further consideration is the fact that sometimes, during the unseen domain transfer, MetalGAN tends to apply unwanted features to the images. For example in `Bushy Eyebrows' the network often changes the hair color to black or applyes mustaches. This is due to the fact that people with bushy eyebrows generally have darker hair and facial hair. The same reasoning can be applyed to `Gray Hair', where the network tends to produce older people, because people with gray hair are usually old.
The reason for this behavior is that, because of the lack of labels, the network has to infer which is the domain to be transferred without any help. This is also a big advantage, because produces a much greater flexibility to the network and allows to add new domains to the network very easily.

\begin{figure}[h]
	\centering
	\includegraphics[width=0.7\textwidth]{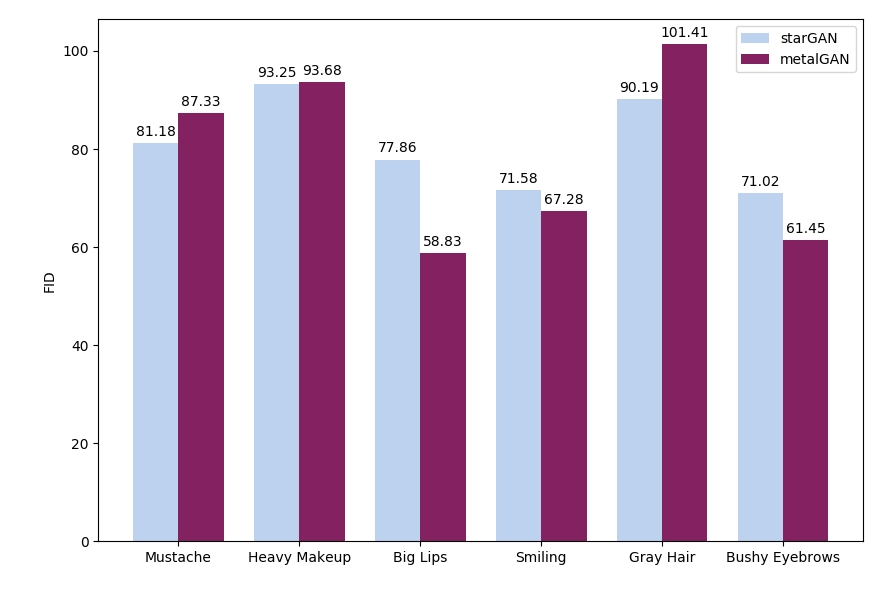}
	\caption{FID score on inference domains (the lower the better).}
	\label{fig:Inference_graph}
\end{figure}
We calculated both FID and PRD also for inference domains, as in the previous section.
In Figure~\ref{fig:Inference_graph}, FID scores are reported. As for training, FID scores depend heavily on the selected domain, but in general, StarGAN and MetalGAN scores are close to each other. In particular, MetalGAN performs better on `Big Lips', `Smiling', and `Bushy Eyebrows', confirming visual evaluation of results.

\begin{figure}[h]
	\centering
	\begin{subfigure}{0.25\textwidth}
		\includegraphics[width=\linewidth]{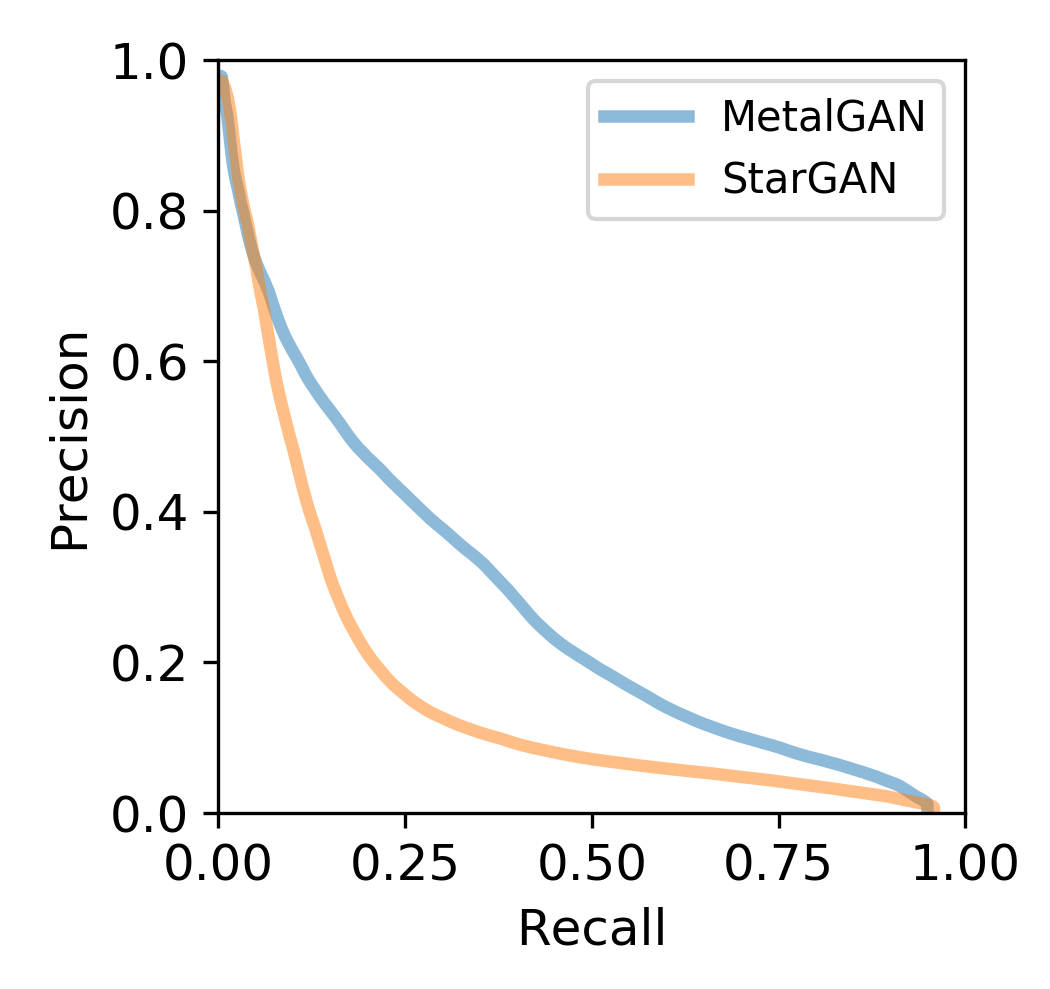}
		\caption{Big Lips}
	\end{subfigure}%
	\begin{subfigure}{0.25\textwidth}
		\includegraphics[width=\linewidth]{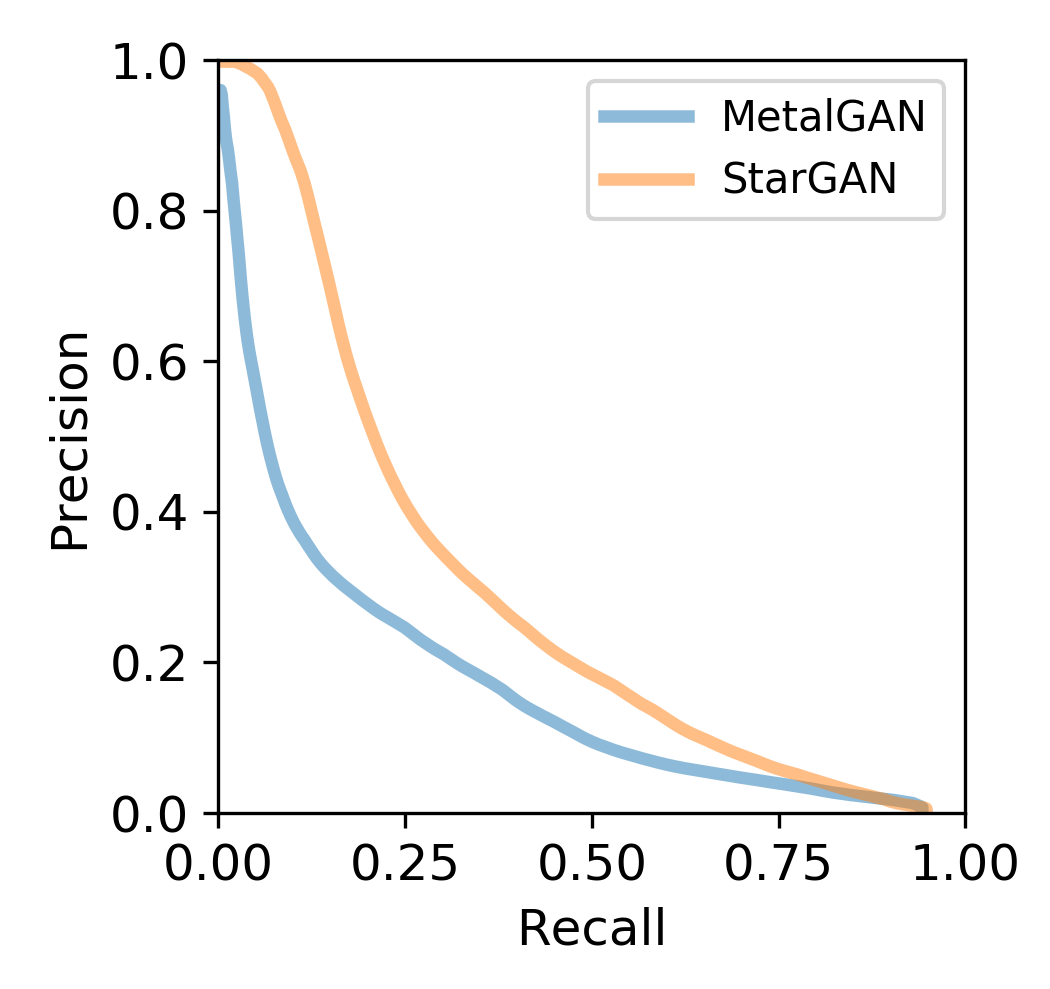}
		\caption{Bushy Eyebrows}
	\end{subfigure}%
	\begin{subfigure}{0.25\textwidth}
		\includegraphics[width=\linewidth]{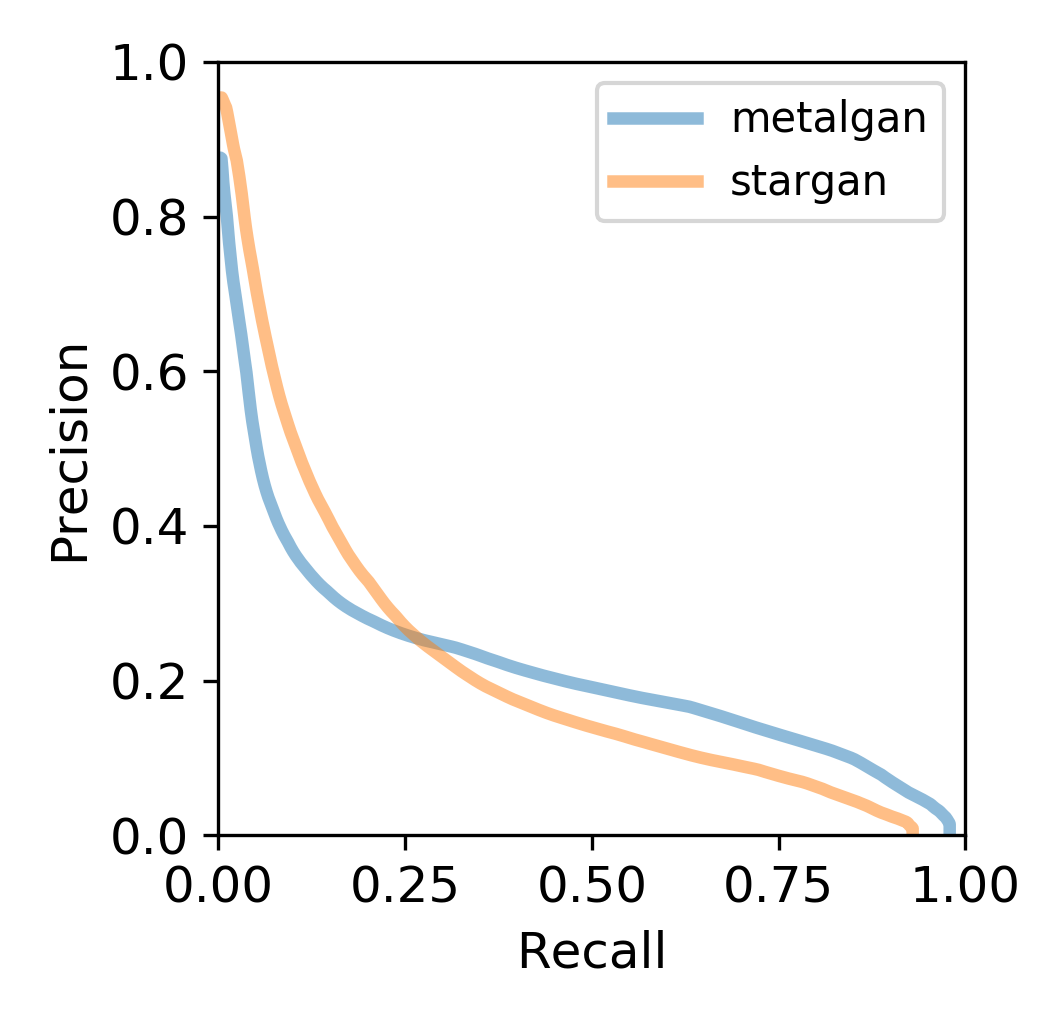}
		\caption{Heavy Makeup}
	\end{subfigure}\\%
	\begin{subfigure}{0.25\textwidth}
		\includegraphics[width=\linewidth]{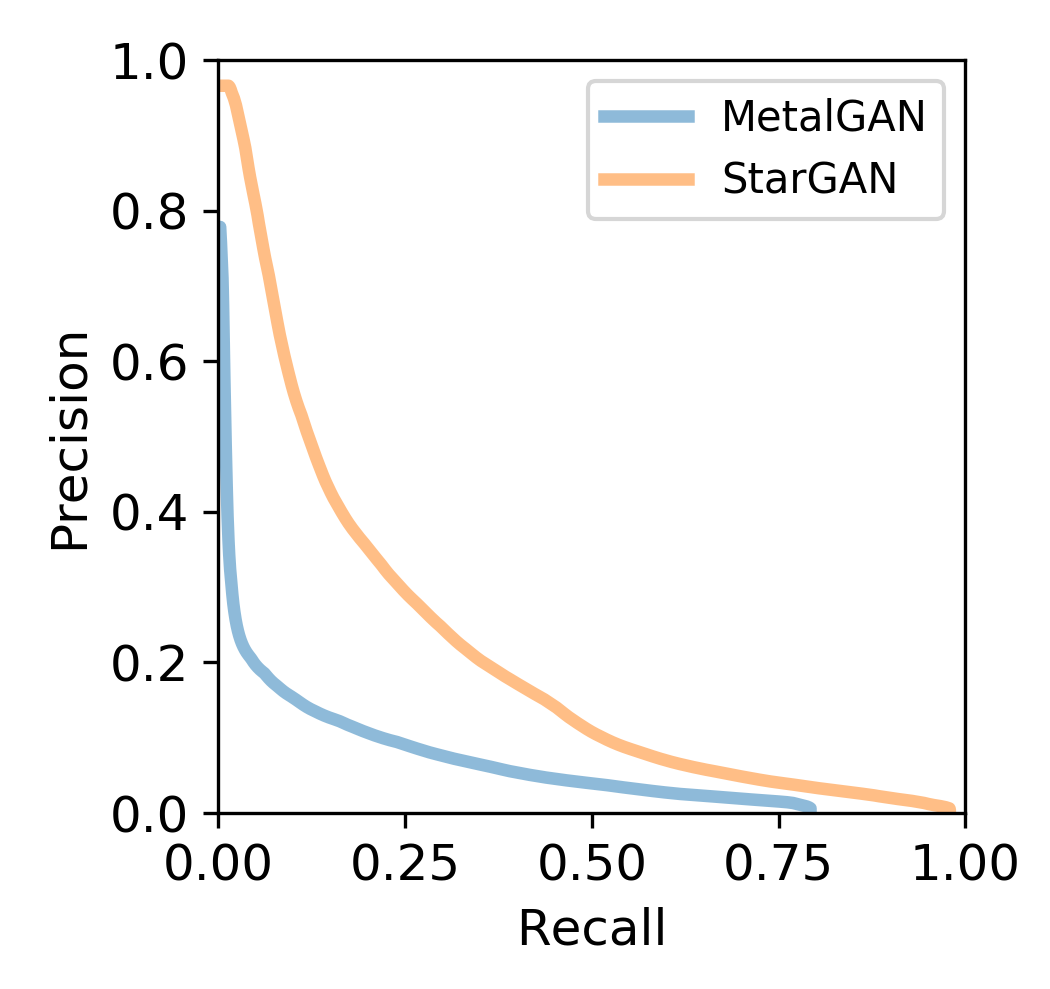}
		\caption{Smiling}
	\end{subfigure}%
	\begin{subfigure}{0.25\textwidth}
		\includegraphics[width=\linewidth]{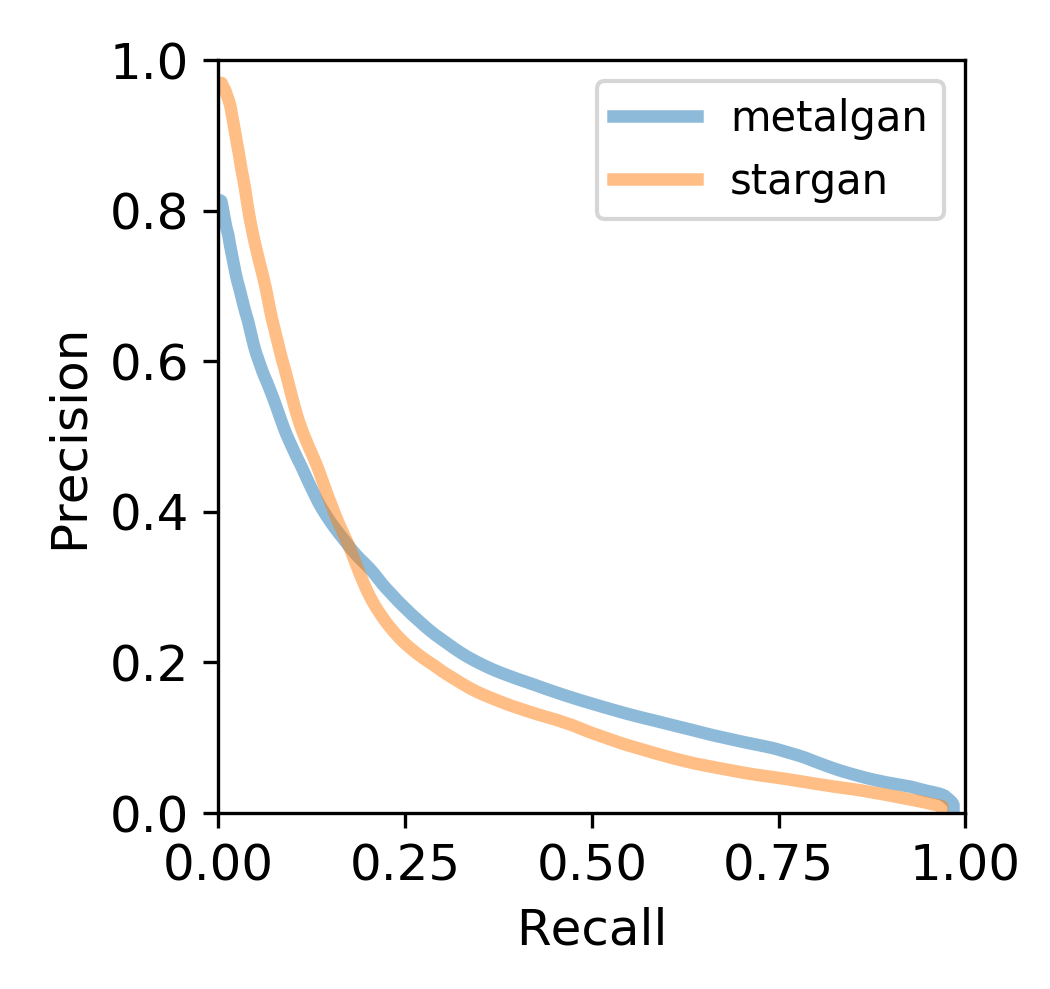}
		\caption{Gray Hair}
	\end{subfigure}%
	\begin{subfigure}{0.25\textwidth}
		\includegraphics[width=\linewidth]{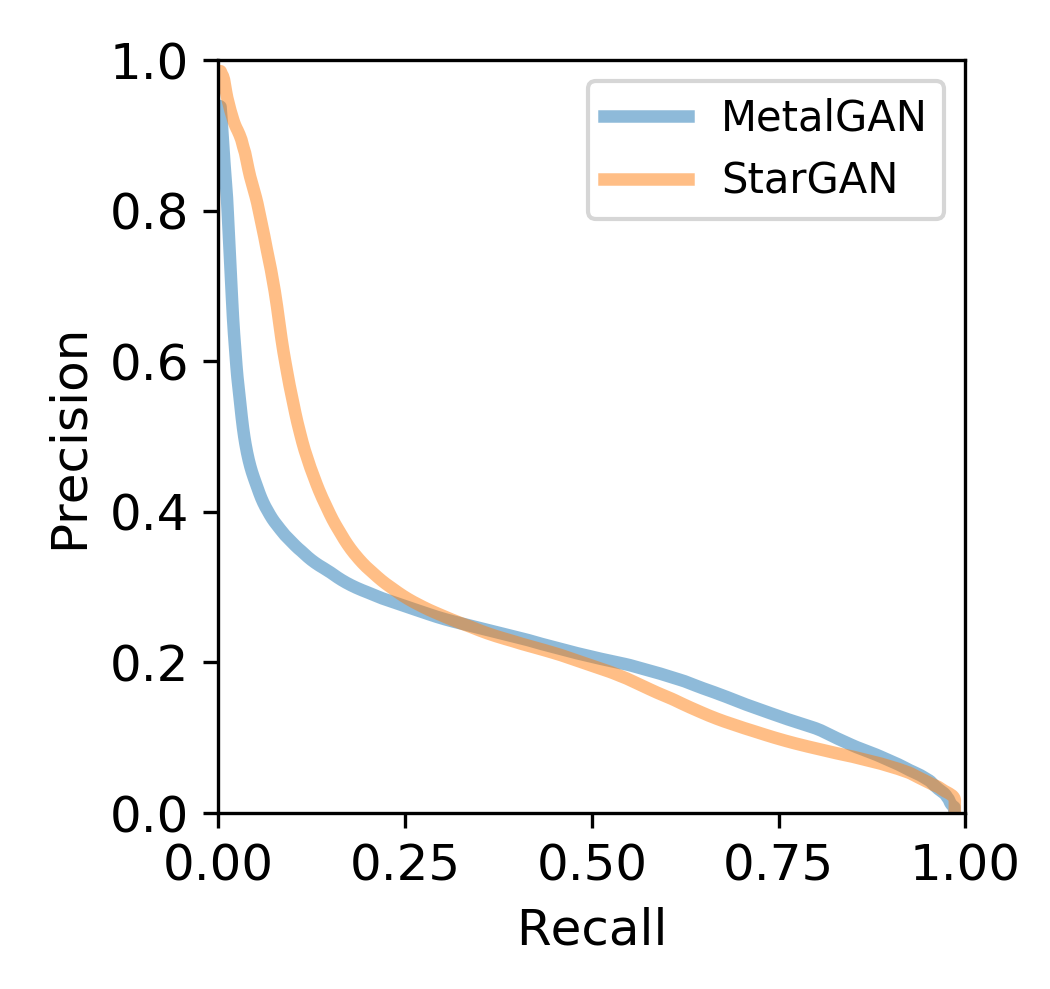}
		\caption{Mustache}
	\end{subfigure}
	\caption{PRD graphs on inference domains.}
	\label{fig:prd_inference_graph}
\end{figure}

\begin{figure}[h]
	\centering
	\includegraphics[width=0.25\textwidth]{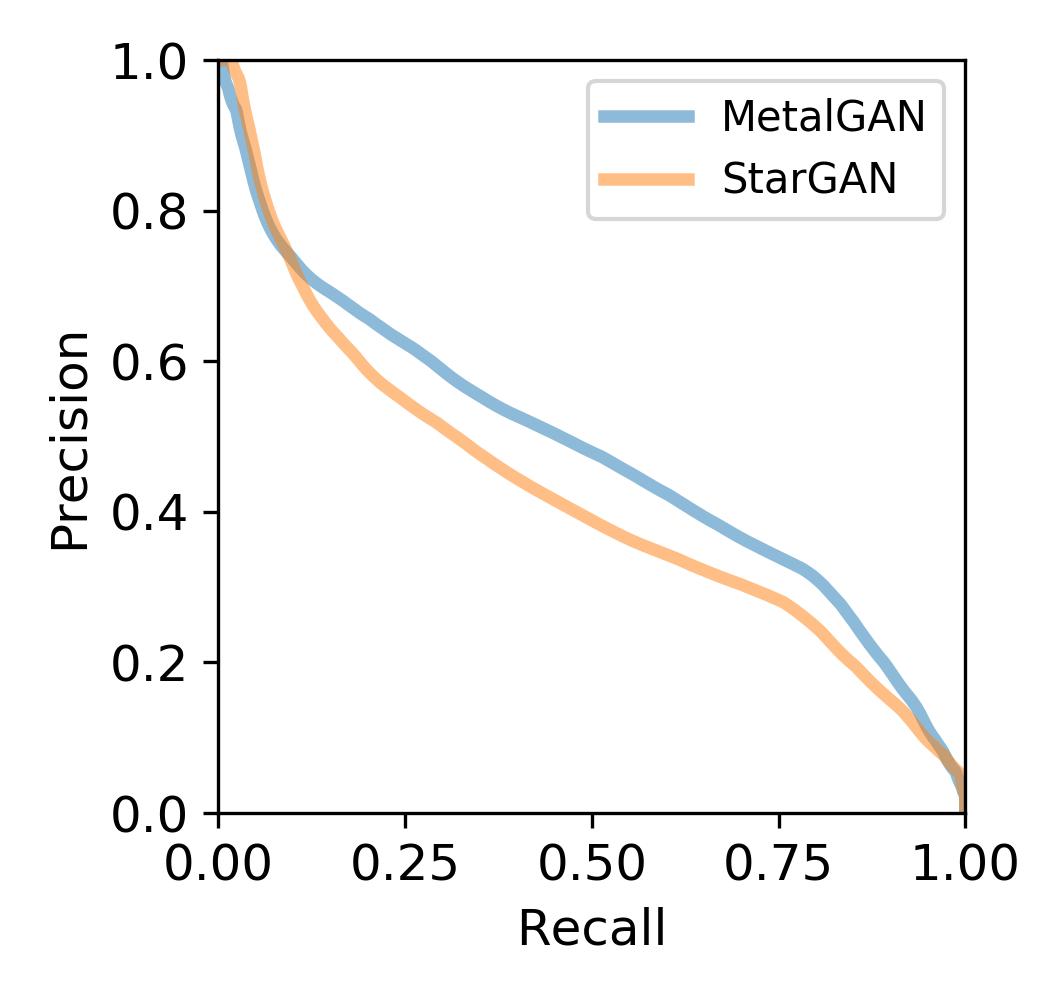}
	\caption{Global PRD on inference domains.}
	\label{fig:prd_allinference_graph}
\end{figure}

In Figure~\ref{fig:prd_inference_graph}, PRD graphs for each unseen domain are reported. All results show how both StarGAN and MetalGAN decrease their precision in this phase, as reasonable. As we can see in Figure~\ref{fig:biglips_inference},~\ref{fig:heavymakeup_inference}, and~\ref{fig:mustache_inference}, the overall quality of the reconstruction is slightly worse than the one of trained domains.
However, despite the unfair comparison, PRD for MetalGAN and StarGAN are pretty similar. Looking at the global PRD, calculated on all six unseen domains at once (Figure~\ref{fig:prd_allinference_graph}), MetalGAN shows better performances especially on distribution recall.

\subsection{Additional Experiments}

\subsubsection{Results on Radboud Faces Database}

\begin{figure}[H]
	\begin{subfigure}[t]{.53\linewidth}
		\centering
		\includegraphics[height=0.75\textheight]{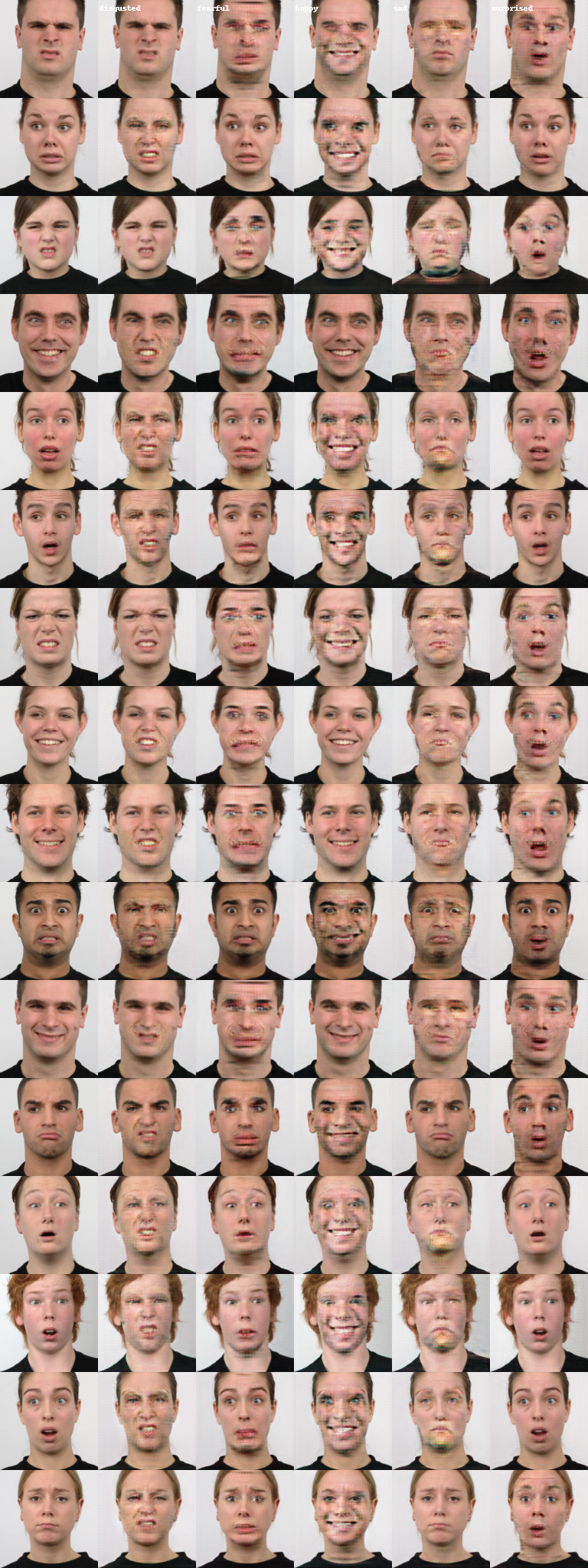}
		\caption{Training domains.}
	\end{subfigure}
	\begin{subfigure}[t]{.3\linewidth}
		\centering
		\includegraphics[height=0.75\textheight]{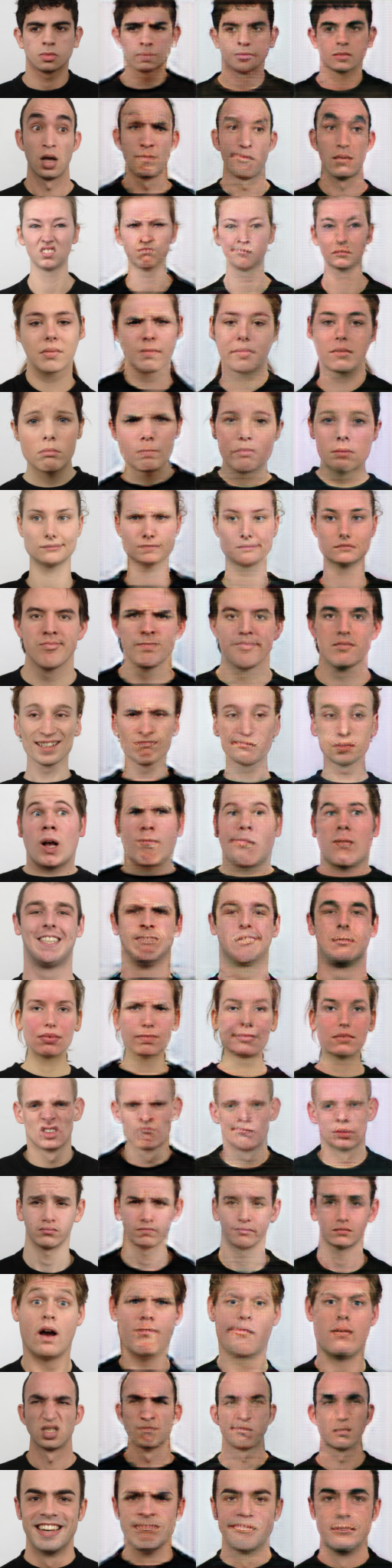}
		\caption{Unseen domains.}
	\end{subfigure}
	\caption{Results on RAFD on trained and unseen domains. Columns in first image represent disgusted, fearful, happy, sad, and surprised domains. Columns in second image represent angry, contemptuous, and neutral domains.}
	\label{fig:rafd}
\end{figure}

In order to further prove the effectiveness of our method, we also trained the $G$-$D$ network on another multi-domain dataset, with MetalGAN algorithm.
Such a dataset is Radboud Faces Database (RAFD) \citep{langner2010presentation}.
RAFD is a set of pictures of 67 models displaying 8 emotional expressions.
We trained the model for 20k iterations with MetalGAN on 5 different emotions (\textit{disgusted}, \textit{fearful}, \textit{happy}, \textit{sad} and \textit{surprised}), maintaining the same configuration used with the CelebA dataset.
A batch of visual results is shown in Figure \ref{fig:rafd} (a), where only the trained domains are tested. Then, additional unseen domains were added to further test our method on this new dataset, as for CelebA. Such new domains are \emph{angry}, \emph{contemptuous} and \emph{neutral}. Inference configurations are the same of Table \ref{tab:inference_exp_settings}. In Figure \ref{fig:rafd} (b), the visual results of MetalGAN inference on RAFD are reported. Results are comparable to trained ones, even if they were obtained by few iterations on the trained model, and with few input images. The naive reason could be that changing the facial expression involves few attributes of the image, thus switching from the input facial expression domain to an unseen one shares a lot of knowledge with the switching between the input and the trained domains. In other words, the main task is the same: changing the facial expression, and little differences between domains are handled easily by the inference steps.

In addition to the qualitative comparison, we also trained a classification network in order to obtain a quantitative evaluation of our method. We choose ResNet-18 as classification network (following the StarGAN paper) and we produced classification results on the different emotions both for our architecture as well as for StarGAN trained on the same domains. Results can be seen in Table \ref{tab:rafd_class}. Following the considerations that were made for the CelebA results, our results for the RAFD dataset are in line with the StarGAN ones, but without the use of label or supervision.
The only exception is the \textit{sad} domain where our network tends to only change the mouth leaving the rest of the face almost unchanged. Therefore, if the input image has another emotion strongly characterized by the eyes or by the eyebrows (such as \textit{surprised}), such features are not changed during the domain switch leading to misclassification.

\begin{table}[h]
\centering
\caption{Classification results on RAFD dataset.}
\begin{tabular}{|l|l|l|l|l|l|}
\hline
 & disgusted & fearful & happy & sad & surprised \\
 \hline
StarGAN & 98.6\% & 97.2\% & 98.6\% & 97.7\% & 97.1\% \\
 \hline
\textbf{MetalGAN (ours)}& 98.4\% & 93.1\% & 97.3\% & 69.7\% & 95.2\%\\
 \hline
\end{tabular}
\label{tab:rafd_class}
\end{table}

\begin{figure}[H]
	\centering
	\includegraphics[width=0.45\textwidth]{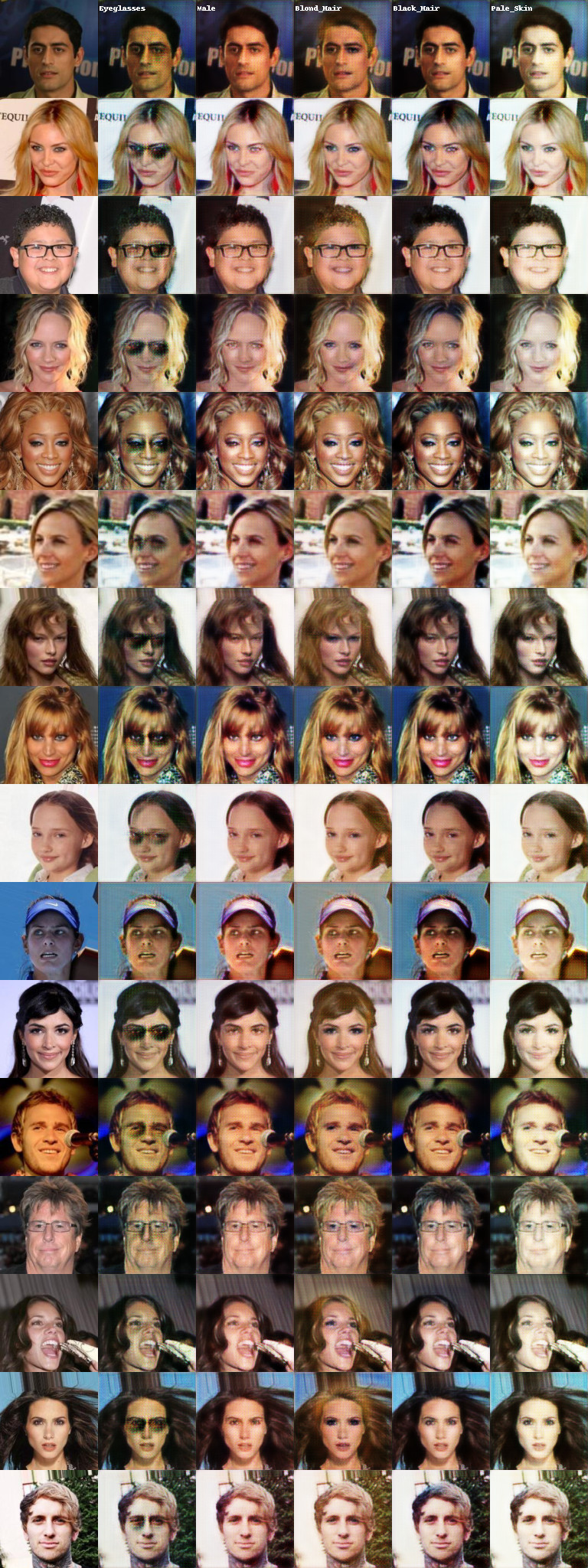}
	\caption{Results on CelebA without the contribution of Domain Loss.}
	\label{fig:no_doloss}
\end{figure}

\subsubsection{Experiments without Domain Loss}

The necessity of the Domain Loss in the MetalGAN algorithm is not self-evident: for this reason, we performed an experiment with the same configurations of MetalGAN standard training (see Table \ref{tab:exp_settings}), but nullifying the domain weight, i.e. $w_{\mathrm{dom}} = 0$.
It is worth noting that this setting still relies on meta-learning, that is the major boost for domain adaptation without labels.
Nevertheless, visual results on CelebA, for MetalGAN without Domain Loss, show that the domain change loses quality, as it should be seen in Figure \ref{fig:no_doloss}.

\section{Conclusions}\label{sec:conclusions}

We proposed a new architecture for multi-domain label-less image-to-image translation. Our system has many features that distinguish it from the state-of-the-art.

First of all, instead of relying on labels for switching the domains, like other state-of-the-art architectures, we had chosen to use meta-learning and in particular Reptile. 
Furthermore, getting rid of labels allowed the architecture to be more flexible, since there is no need of providing hard-coded vectors of labels. It is possible to arbitrarily change the number of domains, and to add a new one during inference. Such an approach was completely unfeasible in previous algorithms like StarGAN, that needs hard-coded labels at training time, and it was a very serious limitation.
Finally, beside the lack of labels, an immediate advantage of the meta-learning approach is that such a method has been used for few-shot learning. Not only, as highlighted above, a new, unseen, task can be added, but in order to do so, just few examples are needed, and neither tedious and long-lasting annotations of labels, nor a full retraining of the model are required.

We proved the effectiveness of our approach with face attributes transfer using the CelebA dataset, and we evaluated it using both FID and PRD metrics. Moreover, we performed additional experiments on RAFD dataset, and tested our approach by nullifying the contribution of Domain Loss, showing its necessity.

Regarding future work, our first objective would be to explore more deeply the possibilities and limitations of meta-learning in order to further improve our algorithm and to prove its effectiveness on others tasks like image generation and semantic segmentation.

\bibliographystyle{elsarticle-harv} 
\bibliography{metalgan}

\end{document}